\documentclass[lettersize,journal]{IEEEtran}
\usepackage{amsmath,amsfonts}
\usepackage{algorithmic}
\usepackage{algorithm}
\usepackage{array}
\usepackage{textcomp}
\usepackage{stfloats}
\usepackage{url}
\usepackage{verbatim}
\usepackage{graphicx}
\usepackage{cite}
\usepackage{soul, color, xcolor}
\usepackage{booktabs}
\usepackage{multirow}
\usepackage[mathscr]{eucal}
\usepackage[hidelinks]{hyperref}

\ifCLASSOPTIONcompsoc
  \usepackage[caption=false,font=normalsize,labelfont=sf,textfont=sf]{subfig}
\else
  \usepackage[caption=false,font=footnotesize]{subfig}
\fi

\begin{document}

\title{An Evolving Scenario Generation Method based on Dual-modal Driver Model Trained by Multi-Agent Reinforcement Learning}

\author{Xinzheng Wu, Junyi Chen, Shaolingfeng Ye, Wei Jiang and Yong Shen
\thanks{This work was supported by the National Key R\&D Program of China under Grant 2024YFB2505705, the National Natural Science Foundation of China under Grant 52232015. (\textit{Corresponding author: Junyi Chen})}
\thanks{All authors are with the School of Automitive Studies, Tongji University, Shanghai 201804, China (e-mail: chenjunyi@tongji.edu.cn).}
}

\markboth{Journal of \LaTeX\ Class Files,~Vol.~14, No.~8, August~2025}%
{Wu \MakeLowercase{\textit{et al.}}: An Evolving Scenario Generation Method based on Dual-modal Driver Model Trained by Multi-Agent Reinforcement Learning}

\IEEEpubid{0000--0000/00\$00.00~\copyright~2025 IEEE}

\maketitle

\begin{abstract}
In the autonomous driving testing methods based on evolving scenarios, the construction method of the driver model, which determines the driving maneuvers of background vehicles (BVs) in the scenario, plays a critical role in generating safety-critical scenarios. In particular, the cooperative adversarial driving characteristics between BVs can contribute to the efficient generation of safety-critical scenarios with high testing value. In this paper, a multi-agent reinforcement learning (MARL) method is used to train and generate a dual-modal driver model (Dual-DM) with non-adversarial and adversarial driving modalities. The model is then connected to a continuous simulated traffic environment to generate complex, diverse and strong interactive safety-critical scenarios through evolving scenario generation method. After that, the generated evolving scenarios are evaluated in terms of fidelity, test efficiency, complexity and diversity. Results show that without performance degradation in scenario fidelity (\textgreater85\% similarity to real-world scenarios) and complexity (complexity metric: 0.45, +32.35\% and +12.5\% over two baselines), Dual-DM achieves a substantial enhancement in the efficiency of generating safety-critical scenarios (efficiency metric: 0.86, +195\% over two baselines). Furthermore, statistical analysis and case studies demonstrate the diversity of safety-critical evolving scenarios generated by Dual-DM in terms of the adversarial interaction patterns. Therefore, Dual-DM can greatly improve the performance of the generation of safety-critical scenarios through evolving scenario generation method.
\end{abstract}

\begin{IEEEkeywords}
Autonomous Driving, Test Scenario, Multi-Agent Reinforcement Learning, Driver Model
\end{IEEEkeywords}

\section{Introduction}
\IEEEPARstart{A}UTONOMOUS Vehicles (AVs) are complex systems composed of multiple integrated subsystems, including environmental perception, decision-making and planning, and control execution systems. Among the various systems involved, the decision-making and planning system serves as the "brain" of the AV's workflow. Upon receiving processed environmental sensing data, this system is responsible for making informed decisions and formulating appropriate plans to guide the control system's actions, thereby playing a pivotal role in the overall operation of the AVs. Particularly for high-level autonomous driving systems (L3-L5), where the system itself serves as the primary entity for Object and Event Detection and Response (OEDR) \cite{SAEJ3016}, any deficiencies in the decision-making and planning system can lead to inappropriate vehicle behaviors when operating in complex traffic environments, potentially resulting in hazardous situations \cite{liu2021crash}. Consequently, ensuring the safety of vehicle decision-making and planning remains a critical challenge that needs to be addressed urgently \cite{li2022decision}.

Currently, scenario-based testing method is an important way to test the safety of decision-making and planning system \cite{zhang2023findinga}. The scenario is an overall dynamic description of the AV along with the roads, other vehicles and other elements in AV's operating environment over a certain period of time \cite{ISO34501}. By generating complex, diverse, and strong interactive safety-critical scenarios, it is possible to simulate the various operating environmental conditions of an AV, and thus obtain the safety performance of the decision-making and planning system.

According to the generation method of testing scenarios, scenario-based testing methods can be classified into testing based on predefined scenarios and testing based on evolving scenarios. For the methods using predefined scenarios, a logical scenario is typically defined based on the three level of scenario abstraction theory \cite{menzel2018scenarios}. Subsequently, concrete testing scenarios are generated by sampling within the parameter space of the logical scenario \cite{wu2025make}. In this type of testing scenario, the behavior of the background vehicles (BVs) is predefined before the testing begins. As a result, BVs can only follow the "script" to complete the relevant actions (for example, changing lanes to the left at the 2nd second of the scenario, with the lane change lasting 3 seconds), instead of interacting with the subject vehicle (SV) controlled by the system under test (SUT).

As a comparison, for the methods using evolving scenario, a continuous traffic environment is created and BVs in the environment are controlled by driver models equipped with decision-making capabilities for testing. Since all vehicles in the environment possess decision-making capabilities, the scenario can continuously evolve into new states based on the bidirectional interaction between the SV and BVs \cite{ma2024evolving}. Furthermore, as the evolution direction of scenarios is directly influenced by the driving style of the driver models controlling the BVs, the selection of adversarial driver models can guide the test scenarios to be safety-critical. Therefore, this kind of methods enables the generation of diverse and highly interactive scenarios that meet the requirements for safety testing of decision-making and planning systems.\IEEEpubidadjcol

\begin{figure*}[b] 
      \centering
      \includegraphics[width=\linewidth]{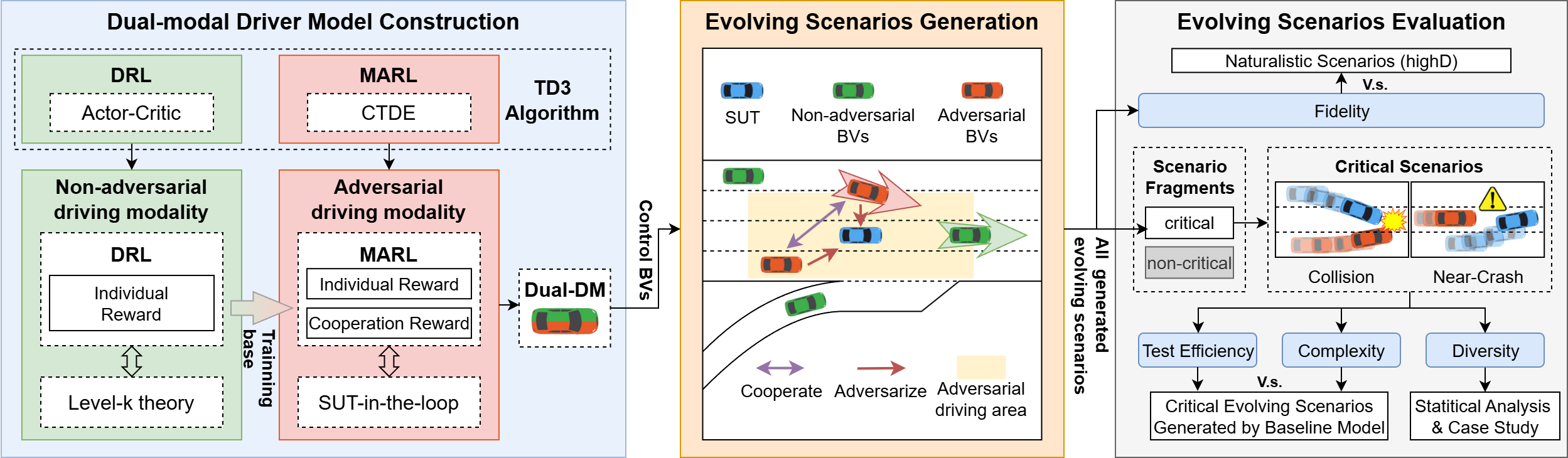}
      \caption{The framework of this paper.}
      \label{framework}
\end{figure*}

Following the idea of generating evolving scenarios, existing studies have developed spatiotemporally continuous testing scenarios \cite{feng2021intelligenta} or self-evolution testing scenarios \cite{ma2024evolving} to evaluate the safety performance of decision-making and planning systems. However, how to balance the behavioral fidelity and adversariality of driver models in evolving scenario generation remains an open question. Overemphasis on adversariality tends to generate irrational collision behaviors (e.g., BVs directly colliding with the SV from all directions). Conversely, overemphasis on fidelity makes the generated evolving scenarios indistinguishable from naturalistic driving data (NDD), resulting in very low safety-critical scenario generation efficiency. Meanwhile, relying solely on single-vehicle adversarial behaviors cannot produce complex and diverse safety-critical scenarios. In light of this, to design and generate evolving scenarios with high testing fidelity, complexity and efficiency, an evolving scenario generation method based on driver model trained by Multi-Agent Reinforcement Learning (MARL) is studied. Through MARL, a dual-modal driver model (Dual-DM) with non-adversarial and adversarial driving modalities is trained and generated. The model is then connected to a continuous simulated traffic environment to generate complex, diverse and strong interactive safety-critical scenarios for the decision-making and planning system efficiently. The framework of this paper is shown in Fig. \ref{framework} and the main contributions are listed as follows:

\begin{itemize}
  \item A dual-modal driver model (namely Dual-DM) with non-adversarial and adversarial driving modalities is designed and trained based on MARL. The model is capable of making adversarial interactions with the SUT through single-vehicle behaviors or multi-vehicle cooperative behaviors.
  \item Evolving scenarios are generated based on the established Dual-DM and safety-critical evolving scenarios for the SUT are filtered.
  \item An evaluation framework for evolving scenarios are proposed. The fidelity, test efficiency, complexity and diversity of the Dual-DM-generated evolving scenarios are verified through comparisons with naturalistic scenarios, evolving scenarios generated by baseline models, as well as statistical analysis and case studies.
\end{itemize}

The remainder of this paper is structured as follows: Section \uppercase\expandafter{\romannumeral2} introduces and analyzes the related works. In Section \uppercase\expandafter{\romannumeral3}, the methodology, including the design and training of Dual-DM, is presented. Section \uppercase\expandafter{\romannumeral4} demonstrates the experimental settings and puts Dual-DM into application. And the generated evolving scenarios are evaluated in Section \uppercase\expandafter{\romannumeral5}. Finally, Section \uppercase\expandafter{\romannumeral6} concludes this paper.

\section{Related Works} \label{sec2}
\subsection{Generation of Evolving Scenarios}
As discussed above, the generation of evolving scenarios are closely related to the driver model that controls the BVs. Using reasonable approaches to build the driver model can improve the generation results of evolving scenarios effectively. In the current literature, the driver models used to generate evolving scenarios can be mainly categorized into function-based models and learning-based models.

Function-based models typically use physical formulas or utility functions to determine the longitudinal and lateral actions of a vehicle. Typical representatives of such models include Intelligent Driver Model (IDM) \cite{treiber2000congested}, MOBIL model \cite{kesting2007general}, and Nilsson model \cite{nilsson2016if}, etc. Function-based models have clear parameters and strong interpretability of vehicle behavior. However, this type of models has many simplified assumptions, which leads to a simple vehicle behavior and makes it difficult to generate complex evolving scenarios that satisfy the safety testing requirements of high-level decision-making and planning systems.

In learning-based models, approaches such as imitation learning (IL) and reinforcement learning (RL) are usually used to build a driver model. IL can build a driver model with high fidelity by learning naturalistic driving data. For example, Chen \MakeLowercase{\textit{et al.}} \cite{chen2019deep} utilized a deep imitation learning method to train and generate an urban driving strategy model, which can generate urban driving scenarios for testing . Tian \MakeLowercase{\textit{et al.}} \cite{tian2023personalized} proposed a personalized planning and control approach for lane change assistance system via end-to-end imitation learning from a few data demonstrations. Sun \MakeLowercase{\textit{et al.}} \cite{sun2024learning} developed a two-dimensional merging behavior model based on a imitation learning framework to generate realistic vehicle trajectories and traffic characteristics. However, such methods have strong data dependence, and the generated models have a poor versatility in different scenarios.

When it comes to RL, Li \MakeLowercase{\textit{et al.}} \cite{li2018game} trained a game theoretic traffic model with longitudinal and lateral decision-making capabilities through RL, and used it to continuously test the decision-making and planning system on straight roads. Wang \MakeLowercase{\textit{et al.}} \cite{wang2022harmonious} introduced a harmonious lane changing strategy based on deep reinforcement learning (DRL), which was generated through convolutional neural network training and could take optimal lane changing actions in the traffic environment by adaptively choosing harmony coefficients. Li \MakeLowercase{\textit{et al.}} \cite{li2022decision} combined DRL with risk assessment function, in order to build a driver model which could find the best lane changing strategy with minimal expected risk. Although RL has difficulties in building a driver model with high fidelity due to the absence of the guidance of data, it can get rid of the limitations of data and generate driver models serving different testing purposes through various settings of reward function. Such models are highly evolving and universally applicable.

In summary, compared to function-based models, driver models constructed via machine learning exhibit greater adaptability in capturing stochastic human behaviors, emulate a wider spectrum of realistic driving patterns, and demonstrate enhanced testability through scenario-based validation frameworks. Among the learning-based models, driver models generated by RL are more versatile and flexible in different scenarios. Therefore, further research on how to generate a driver model based on RL, which can contribute to the improvement of the ability of generating safety-critical scenarios, is demanded.

\begin{figure}[b] 
      \centering
      \includegraphics[width=\linewidth]{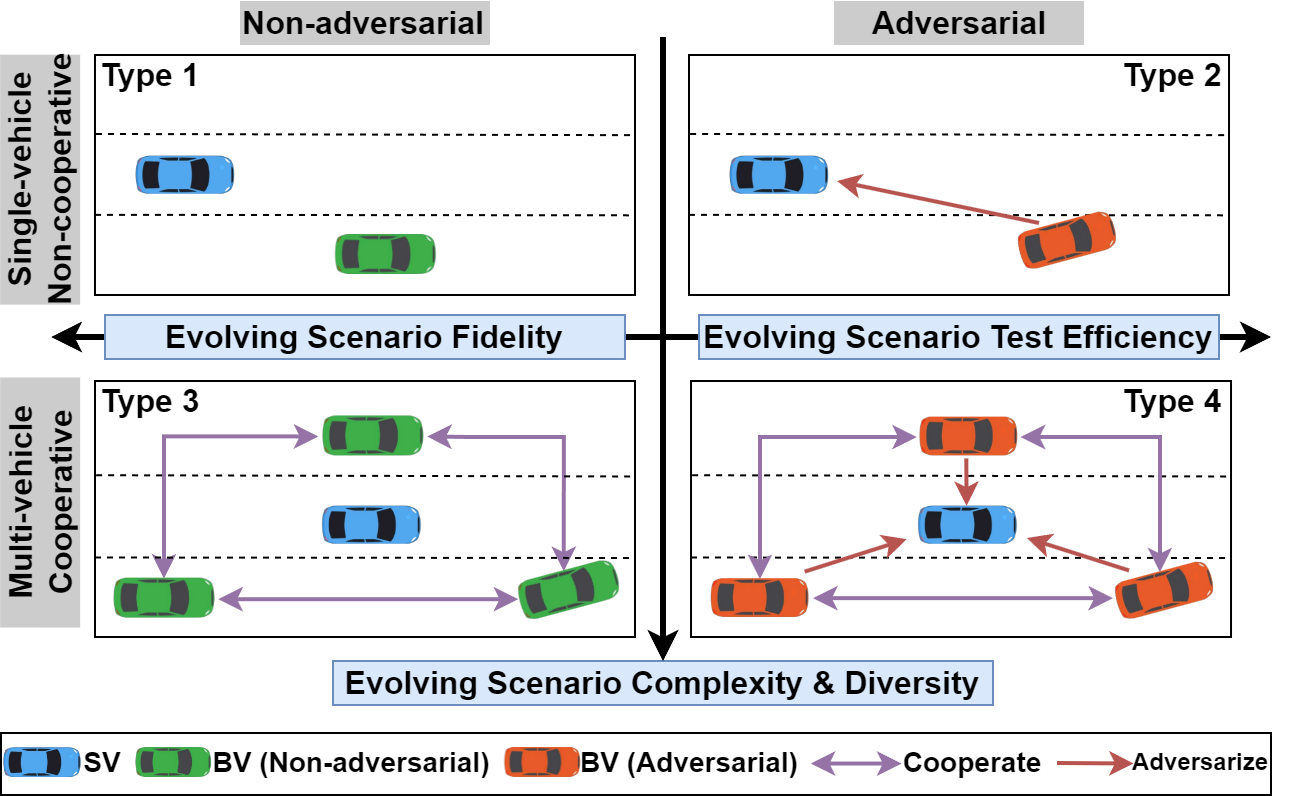}
      \caption{Different driving characteristics of driver models based on RL.}
      \label{DriverModelClass}
\end{figure} 

\subsection{Driver Model Construction Methods based on RL}
The construction method of RL-based driver models can be classified along two dimensions: adversariality of BVs toward the SV and collaboration among BVs. In terms of the adversarial characteristics against the SV, it can be divided into non-adversarial characteristics and adversarial characteristics. In terms of the cooperation characteristics between BVs, it can be divided into single-vehicle non-cooperative characteristics and multi-vehicle cooperative characteristics. As demonstrated in Fig. \ref{DriverModelClass}, evolving scenarios generated by non-adversarial driver models exhibit high fidelity, while those generated by adversarial driver models typically demonstrate high testing efficiency. Furthermore, driver models with multi-vehicle collaboration characteristics can generate complex and diverse evolving scenarios.

In studies on driver model type 1 \cite{zhu2018humanlike, huang2022driving, liao2024modelling}, naturalistic driving characteristics are regarded as the modeling benchmark, and model characteristics are modified differently in terms of driver style \cite{chu2023review} and social value orientation \cite{crosato2023interactionaware}. These models are capable when serving as the decision-making and planning system for the SV, demonstrating human-like or personalized driving behaviors. However, for the goal of generating critical test evolving scenarios in this paper, since these models do not have adversarial settings, when they are used to control BVs in interactive environment, the efficiency of generating critical scenarios is relatively low \cite{jiang2023generation}. 

For the researches on driver model type 2, RL and decision tree can be combined to train a driver model with the ability to challenge the SV, thereby generating scenarios with high testability \cite{feng2021intelligenta}. In addition, key fragments in the data can be extracted and reorganized, allowing the driver model to learn adversarial driving characteristics rapidly, which increases the verification speed of AV by multiple orders of magnitude \cite{feng2023dense}. These models can effectively improve the criticality of the generated scenario. Moreover, by integrating naturalistic distributions of vehicle maneuvers from NDD, the fidelity of these models can also be guaranteed. However, since the adversarial patterns that a single vehicle can generate are limited, the complexity and diversity of the scenarios produced by such models are insufficient.

When it comes to the researches on driver model type 3, there are a large number of studies using MARL to build models, so that the models possess cooperative lane changing \cite{chen2023multiagent, wang2024multiagent, guo2024modeling}, on-ramp merging and other capabilities \cite{chen2023deep}. However, similar to  driver model type 1, the cooperative characteristics of driver model type 3 only serve the driver models’ own driving ability, and the model does not make adversarial behaviors against the SV. Therefore, just like driver model type 1, its testability of the decision-making and planning system is poor.

By contrast, the scenarios generated by driver model type 4 are more complex and testable. Possessing the capability of multi-vehicle cooperation, multiple BVs can gradually compress the SV's drivable space through cooperative strategies, thereby exposing deficiencies in the SV's decision-making and planning. This type of model does not require individual BV to deliberately conflict or collide with the SV, thus enabling efficient generation of complex and diverse safety-critical evolving scenarios while ensuring the behavioral realism of BVs. However, in existing literature, the multi-vehicle joint decision model is less universally applicable in different complex scenarios \cite{he2023adversarial}, and the effectiveness of the multi-vehicle independent decision model in complex testing environments remains unknown \cite{wachi2019failurescenarioa}.

To sum up, following the concept of the fourth type of driver model, this paper aims to train a driver model with collaborative adversarial capabilities based on MARL, thereby efficiently generating complex and diverse critical test evolving scenarios without compromising scenario fidelity.

\section{Dual-modal Driver Model Construction} \label{sec3}
In this section, Dual-DM and its training framework are proposed. Then the specific configurations of the proposed driver model such as observation space, action space, and reward function are designed. Finally, the training of Dual-DM is completed.

\subsection{Design of Dual-DM}
In the evolving scenarios, running vehicles are divided into an SV and BVs according to different roles. Among them, the SV utilizes the decision-making and planning system under test (which is referred to as SUT in this paper) to make behavioral decisions. The BVs use a driver model (which is referred to as  Non-Player Character (NPC) model in this paper) with decision-making and planning capabilities to make behavioral decisions. BVs controlled by the NPC model directly interact with the SV controlled by the SUT. Therefore, the driving behaviors and interaction capabilities of the NPC model directly determine the safety-critical scenario generation performance of the evolving scenarios.

\begin{figure}[b] 
      \centering
      \includegraphics[width=\linewidth]{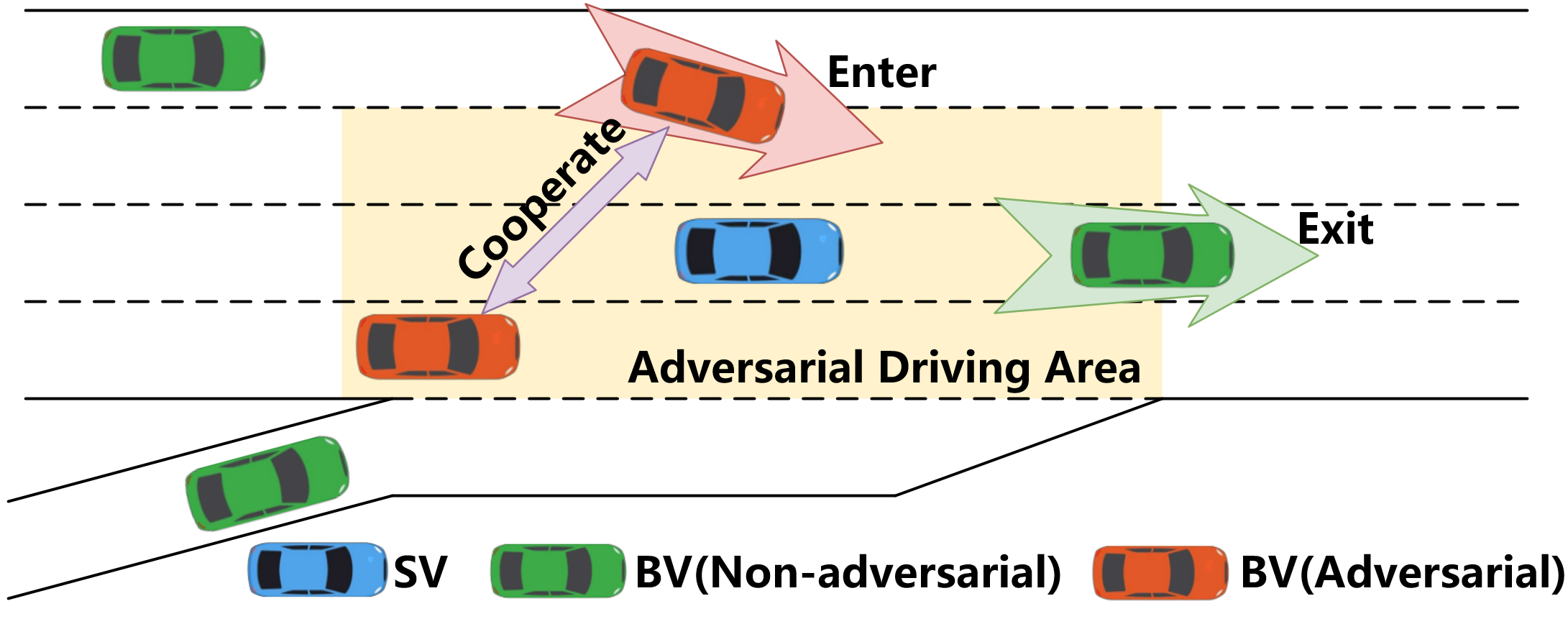}
      \caption{Illustration of driving modality switching of Dual-DM.}
      \label{ModalitySwitch}
\end{figure} 

In this study, Dual-DM is designed to serve as an NPC model in the evolving scenarios to improve the safety-critical scenario generation capabilities. As the name of the proposed driver model suggests, Dual-DM operates with two driving modalities: When the BV controlled by the driver model travels far away from the SV, the model will make decisions according to its own driving goals and drive individually, which is manifested as "non-adversarial driving modality". When the BV travels at a position close to the SV, the driver model will control BV to interact confrontationally with the SV controlled by the SUT through single-vehicle behaviors or multi-vehicle cooperative behaviors, which is manifested as "adversarial driving modality", thereby increasing the probability of the SUT facing safety-critical scenarios. In this study, the area where the driver models convert to the adversarial driving modality is defined as the "adversarial driving area", as shown in the yellow area of Fig. \ref{ModalitySwitch}. Note that in the adversarial driving modality, when there is only one single BV in the adversarial driving area, the BV will directly interact with the SV. When there are multiple BVs in the adversarial driving area, they can confront the SV together through mutual cooperation. Specifically in this paper, the adversarial driving area is set as a 45-meter longitudinal range around the SV (with 22.5 meters in the front and 22.5 meters in the rear), and laterally spans five lanes (two lanes to each side of the SV's current lane).

Moreover, the adversarial behavior designed in this study avoids scenarios where BVs directly collide with the SV. Instead, it employs BVs to progressively compress the SV's drivable space while ensuring a controlled reaction time window for the SUT. In other words, this methodology aims to trigger safety-critical scenarios arising from the SUT's failure to respond appropriately to emergent spatial constraints, thereby evaluating its safety performance. To address this objective, the study concentrates on two critical spatial domains: the drivable space within the SV's current lane and potential lane-changing spaces in adjacent lanes. BVs positioned within the adversarial driving area are required to constrain the SV's forward drivable space through either lane-changing maneuvers or velocity adjustments (acceleration/deceleration). Concurrently, these BVs must optimize their longitudinal positioning relative to the SV in neighboring lanes, thereby effectively restricting potential lane-changing spaces of the SV.

\subsection{Training Framework of Dual-DM}

\begin{figure*}[t] 
      \centering
      \includegraphics[width=\linewidth]{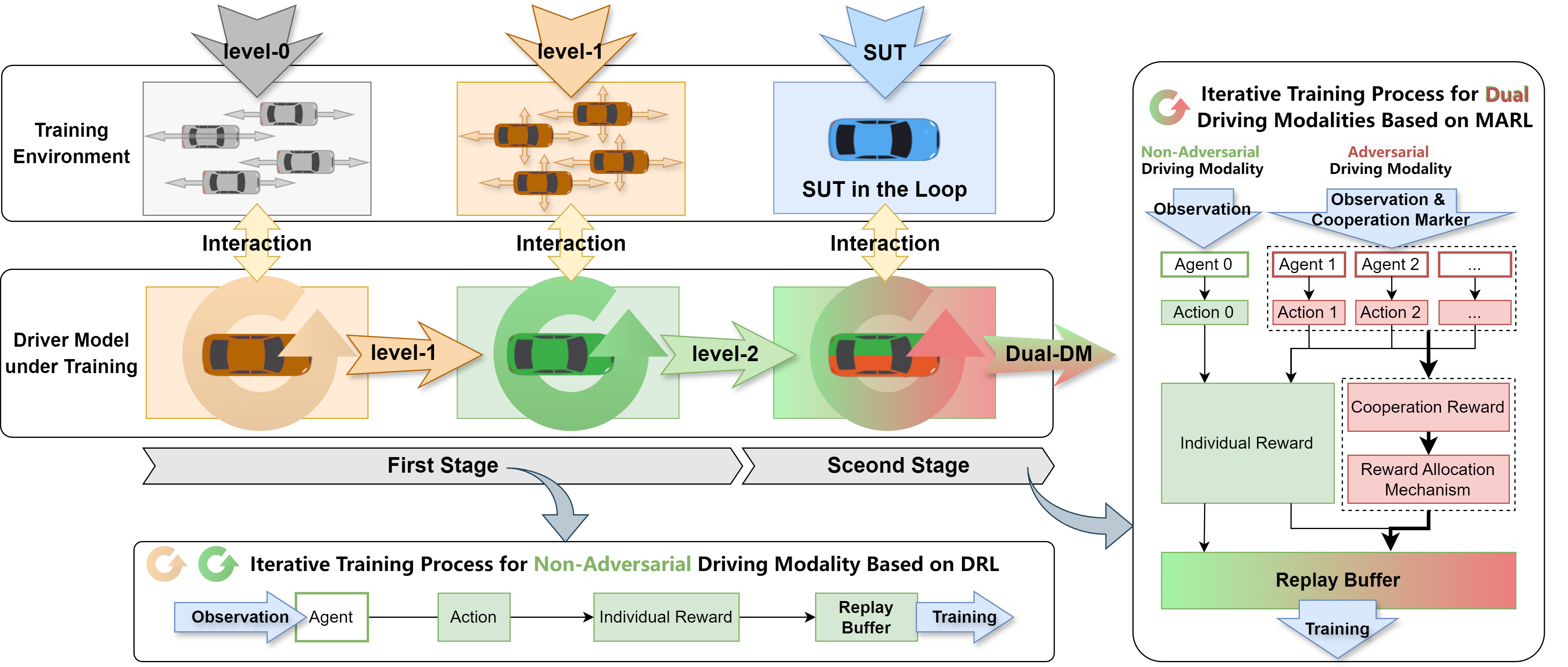}
      \caption{Training Framework of Dual-DM.}
      \label{TrainingFramework}
\end{figure*}

In this study, a two-stage model training framework is designed as shown in Fig. \ref{TrainingFramework}. In the first stage, a driver model training method based on DRL by our previous work \cite{ma2024evolving} is used to equip the driver model with basic single-vehicle driving capabilities. This training method completes the training of the driver model's non-adversarial driving modality based on the DRL algorithm of the Actor-Critic structure, and improves the model's interaction ability through hierarchical training based on level-k theory. 

In the second stage, MARL is used to train the adversarial driving modality of the driver model. This is based on the fact that the driver model has completed the first stage of training on the non-adversarial driving modality, after which the driver model is given the adversarial driving modality in the second stage.

Specifically, the model's adversarial driving modality are trained through a SUT-in-the-loop approach. Given the variance in SUT configurations, different SUTs may exhibit significantly different behaviors when facing the same scenario. Therefore, in order to achieve strong confrontation against a specific SUT, it is necessary to connect the SUT to the training environment and train the driver model through an SUT-in-the-loop approach. The training environment of the second stage contains one vehicle controlled by the SUT and multiple vehicles controlled by the trained driver model. With the policy of SUT unchanged, the driver model dynamically updates and ultimately develops effective adversarial behaviors through iterative adversarial interactions with the SUT. 

Furthermore, MARL based on Centralized Training with Decentralized Execution (CTDE) structure is used for driver model training. In the conventional CTDE structure, cooperation rewards and the Critic networks are shared by \textit{Agents} in the training environment, but the Actor networks are trained separately using data obtained by each of the \textit{Agents} \cite{park2023multiagent}. In such method, the experience replay buffer of each model is limited, so it takes a long time to complete the training of exploration and cooperation capabilities in multiple scenarios. To address this issue, in the training environment of this study, the same Actor network and Critic network are used by all \textit{Agents}, and a shared cooperative reward is applied to complete the training of the model. Such design can guide the driver model to learn cooperative behaviors while making full use of the training data obtained by each \textit{Agent} from exploring the environment. 

The detailed training scheme for dual driving modalities is shown in the right part of Fig. \ref{TrainingFramework}. The driver model employed in the second training stage builds upon the level-2 model from the first stage. During the training process, observation data from all \textit{Agents} is collected to train the model, which aims to train adversarial driving modality while maintain the non-adversarial driving modality that have been already trained. More in detail, for \textit{Agents} outside the adversarial driving area, only individual rewards based on their behaviors are given, which is consistent with the first stage of training. For \textit{Agents} located in the adversarial driving area, in addition to the individual rewards to ensure fundamental driving capabilities, an additional cooperative reward based on the results of the adversarial interaction with the SUT is calculated. The reward is shared among all \textit{Agents} in the adversarial driving area, and is allocated to each \textit{Agent} according to the reward allocation mechanism. Notably, observation data obtained by \textit{Agents} in the adversarial driving area will be assigned a cooperation marker, which is used to distinguish training data of different driving modalities during the training and determine which modality the driver model should display after training.

The training algorithm of both training stages uses the Twin Delayed Deep Deterministic Policy Gradient (TD3) algorithm \cite{fujimoto2018addressing}. In the first stage, the TD3 algorithm is used to train and update the driving strategy of the sole \textit{Agent} in the training environment. While in the second stage, the TD3 algorithm updates the Actor and Critic network shared by all \textit{Agents} as well as the driving strategies of all \textit{Agents} in the environment after each round of training.

\subsection{Dual-DM Formulation}
\subsubsection{Observation Space}

\begin{figure*}[t] 
      \centering
      \includegraphics[width=.9\linewidth]{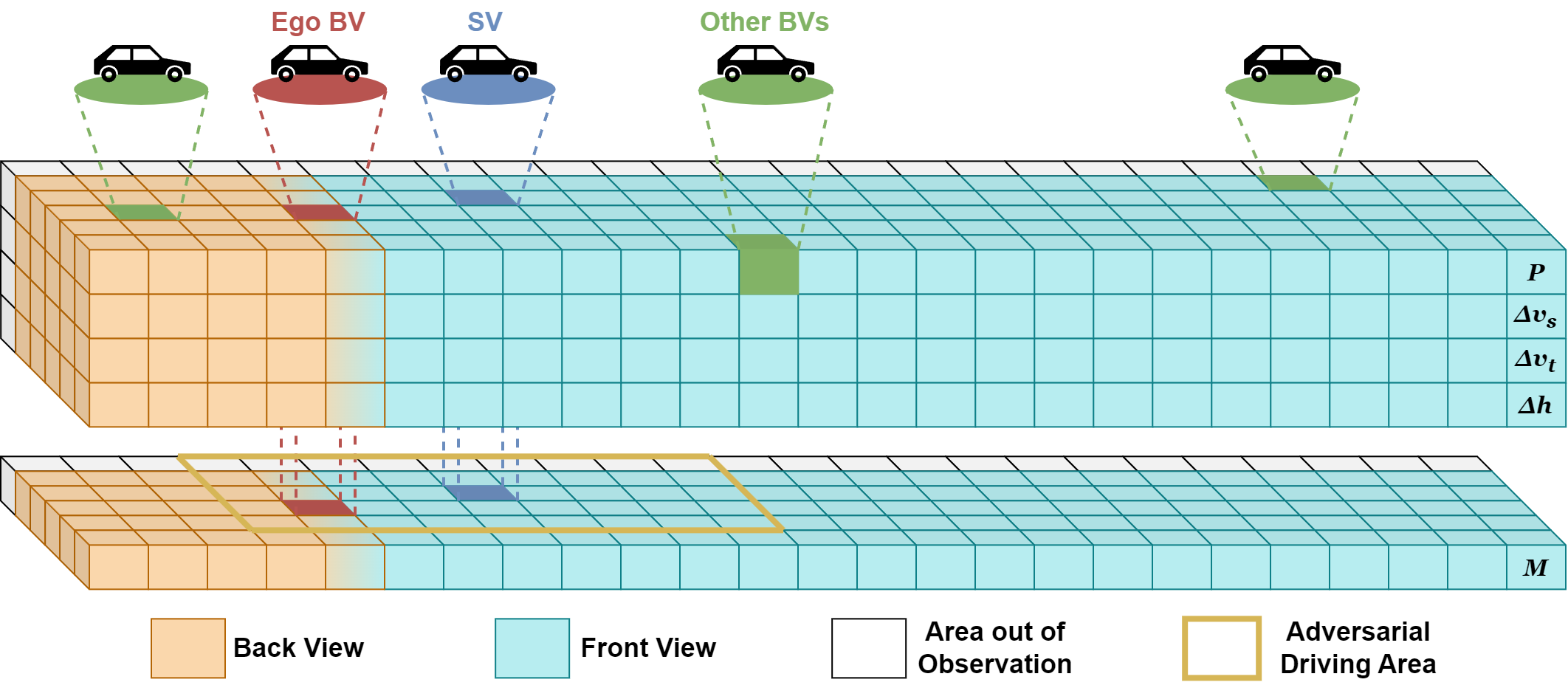}
      \caption{Illustration of the observation space.}
      \label{ObservationSpace}
\end{figure*}

Based on an ego-centric perspective, the driver model monitors road conditions and other vehicles' states within the surrounding environment. The observation scope spans 100 meters ahead and 20 meters behind longitudinally, while laterally encompassing five lanes: the host lane plus two adjacent lanes on both left and right sides. An illustration of the observation space is shown in Fig. \ref{ObservationSpace}.

Specifically, the environmental observation range is discretized by dividing the grid to facilitate the driver model to process the surrounding environment information. The grid is set as a rectangular space of $5\times3.5$ meters, where the grid width is configured as 3.5 meters to align with the lane width in the map. Longitudinally, the 120-meter observation range is divided into 24 rows of grids. Including the grid where the ego vehicle is located, the longitudinal direction comprises a total of 25 rows. Laterally, the 5 observation lanes are divided into 5 columns of grids, resulting in a 25×5 gridded observation space. After acquiring environmental data, the driver model maps the information into a 25×5 observation matrix. 

As shown in Fig. \ref{ObservationSpace}, the observation information obtained by the driver model includes five layers: spatial drivability layer $P$, relative longitudinal velocity layer $\Delta v_s$, relative lateral velocity layer $\Delta v_t$, relative yaw angle layer $\Delta h$, and adversarial driving determination factor layer $M$, which constitutes a $25\times5\times5$ observation space. Among them, space drivability layer $P$ indicates whether the corresponding grid is occupied by vehicles or not. The value is set to 1 if the grid contains obstructing vehicles, otherwise it is set to 0. Notably, only when a gird is occupied by a vehicle, the other 4 parameter layers corresponding to this grid are assigned values to determine the relative relationship between the obstructing vehicle and the ego BV; when no vehicle exists in the grid, the other observation layers are assigned a default value of 0.

Moreover, the adversarial driving determination factor layer $M$ records whether the a grid is occupied by SV. Only when the ego BV is within the adversarial driving area, this layer records the position of the SV by assigning a value of 1 to the corresponding grid, indicating the relative relationship between the ego BV and SV. Otherwise, all values in this layer are initialized to 0. According to the values in this layer, the driving modality of ego BV can be determined.

\subsubsection{Action Space} \label{sec_actionSpace}

The action output of the driver model is a combination of longitudinal acceleration and deceleration actions and lateral lane change actions. Both longitudinal acceleration and deceleration actions $\mathcal{A}_{long}$ and lateral lane change actions $\mathcal{A}_{lat}$ are set to continuous variables within the range of $[-1,1]$.

The longitudinal acceleration and deceleration actions $\mathcal{A}_{long}$ are directly output to the vehicle entity controlled in CARLA as throttle and brake signals. When $\mathcal{A}_{long}>0$, the throttle control signal $\mathcal{U}_{throttle} = \mathcal{A}_{long}$, the brake control signal $\mathcal{U}_{brake}=0$, and the vehicle is in an accelerating state. When $\mathcal{A}_{long}<0$, the brake control signal $\mathcal{U}_{brake}=-\mathcal{A}_{long}$, the throttle control signal $\mathcal{U}_{throttle}=0$, and the vehicle is in a decelerating state.

The lateral lane change action $\mathcal{A}_{lat}$ is discretized and transmitted to the lateral control model as a decision result $\mathcal{D}_{LaneChange}$ of the lateral lane change action, which is then converted into a control signal through the control model and output to the vehicle entity controlled in CARLA. Lane change decisions are discrete variables with values $\mathcal{D}_{LaneChange}\in \{-1,0,1\}$, which represent right lane change, lane keeping, and left lane change respectively. When the lateral lane-changing action $\mathcal{A}_{lat}>0.5$, $\mathcal{D}_{LaneChange}=1$. When the lateral lane-changing action $\mathcal{A}_{lat}<-0.5$, $\mathcal{D}_{LaneChange}=-1$. In other cases, $\mathcal{D}_{LaneChange}=0$.

\subsubsection{Reward Function} \label{sec_Reward}
According to the training framework, the reward function for driver model training is decomposed into two components: individual reward and cooperative reward.

\textbf{Individual reward} is designed based on the non-adversarial driving modality and is used to guide DRL and MARL training so that the driver model has the fundamental driving ability of tracking, car following, lane changing, on-ramp merging, etc. The individual reward is set to:

\begin{equation}
\label{eq1}
    r_{ind} = 
    \begin{cases} 
      r_c, & \text{if collision or boundary violation} \\ 
      r_{cf} + r_{lc} + r_{rv}, & \text{otherwise}
    \end{cases}
\end{equation}

As shown in Eq. \ref{eq1}, when the model causes a collision with other vehicles or drives out of the road boundary, the individual reward is directly awarded a negative reward $r_c$. In other cases, the sum of the three rewards $r_{cf}$, $r_{lc}$, and $r_{rv}$ is calculated as the individual reward result.

$r_{cf}$ is used to guide the model's longitudinal driving behaviors by quantifying both post-action states and state differentials before/after action execution. $r_{cf}$ increases when: 1) The car-following distance approaches the target distance. 2) The post-action following distance is closer to the target distance compared to the pre-action distance. The calculation formula of $r_{cf}$ is:

\begin{equation}
\label{eq2}
    r_{cf} = \mu \times r_{d,{t_1}} + \lambda \times (r_{d,{t_1}} - r_{d,{t_0}})
\end{equation}
where $r_{d,{t_0}}$ and $r_{d,{t_1}}$ are the model's following distance rewards before and after the execution of the action, and $\mu$ and $\lambda$ are the weight coefficients respectively. The model following distance reward $r_d$ quantifies the gap between the model's actual following distance and the target following distance, and can be calculated as follows:

\begin{equation}
\label{eq3}
 r_d = -\frac{(d_{front} - d_{desired})^2}{\delta_1}
\end{equation}
where $d_{front}$ is the distance between the model and the front vehicle, $d_{desired}$ is the desired following distance, and $\delta_1$ is the reward discount coefficient.

$r_{lc}$ is used to guide the lateral lane change behavior of the model. Since the lane change is used to avoid obstacles or gain more drivable space ahead, the greater the forward drivable distance of the target changing lane is, the greater $r_{lc}$ is. The calculation formula of $r_{lc}$ is:

\begin{equation}
    r_{lc} = \mathrm{sgn}(d_{target} - d_{current}) \left(\frac{d_{target} - d_{current}}{\delta_2}\right)^2 + p_{lc}
\label{eq4}
\end{equation}
where $d_{target}$ and $d_{current}$ represent the drivable distance ahead in the target changing lane and current lane, respectively, and $\delta_2$ is the reward discount coefficient. $p_{lc}$ is the penalty factor for lane change behaviors, which gives a small negative reward when the model makes a lane change decision, in order to prevent the model from generating worthless random lane change behaviors.

$r_{rv}$ is used to prevent the model from illegal and irrational driving behaviors. When the model commits illegal or irrational driving behaviors, a negative reward $r_{rv}$ is awarded. The determination for illegal and irrational driving behaviors is elaborated in detail in Section \ref{behavConstraints}.

In this paper, the values of different parameters for individual rewards are set as listed in Table \ref{tab1}.

\begin{table}[b]
\renewcommand{\arraystretch}{1.3}
\caption{Parameter Settings of Individual Reward }
\label{tab1}
\begin{center}
\begin{tabular}{ m{0.5\linewidth}  m{0.16\linewidth}<{\centering} m{0.16\linewidth}<{\centering} }
\toprule
\textbf{Individual reward components} & \textbf{Parameter} & \textbf{Value}  \\
\midrule
Collision reward $r_c$ & $r_c$ &  -3 \\ \hline

\multirow{4}{.8\linewidth}{Longitudinal driving behavior reward $r_{cf}$} & $\mu$ & 1   \\ \cline{2-3}

~ & $\lambda$ & 10 \\ \cline{2-3}

~ & $d_{desired}$ & 15  \\ \cline{2-3}

~ & $\delta_1$ & 8000  \\ \hline

\multirow{2}{\linewidth}{Lateral driving behavior reward $r_{lc}$} & $\delta_2$ & 3000 \\ \cline{2-3}

~ & $p_{lc}$ & -0.1  \\ \hline

Illegal driving behavior reward $r_{rv}$ & $r_{rv}$ & -0.3 \\
\bottomrule
\end{tabular}
\end{center}
\end{table}

\textbf{Cooperative reward} is designed based on adversarial driving modality and is used to guide MARL training. It leads the driver model to compress the space near SV through behavioral cooperation and achieve adversarial interactions against SV. Accordingly, based on the changes in environmental conditions near SV, cooperative reward is given to the driver models in the adversarial driving area, and the reward is set as:

\begin{equation}
    r_{coop} = r_l + r_r + r_f
\label{eq5}
\end{equation}
where $r_l$ and $r_r$ reward the compression of the left and right lane-changing space of the SV respectively, and $r_f$ rewards the compression of the drivable space in the front of the lane where the SV is located. The calculations of $r_l$ and $r_r$ are as follows:

\begin{equation}
r_{i} = \frac{\min \left[ ( d^{f}_{i,{t_0}} )^2, ( d^{b}_{i,t_0} )^2 \right] - \min \left[ ( d^{f}_{i,t_1} )^2, ( d^{b}_{i,t_1} )^2 \right]}{\eta_1}
\label{eq6}
\end{equation}
where $i\in[\text{left},\text{right}]$, $d^{f}_{i,t_0}$ and $d^{b}_{i,t_0}$ represent the longitudinal distances between SV and the nearest vehicles in front and behind on the left/right lane respectively before the driver model performs the action. $d^{f}_{i,t_1}$ and $d^{b}_{i,t_1}$ represent the same values after the driver model performs the action. $\eta_1$ is the reward discount coefficient, which is set to 400 in this paper.

$r_f$ is calculated as follows:

\begin{equation}
r_f = \frac{(d_{t_0}^{f})^2-(d_{t_1}^{f})^2}{\eta_2}
\label{eq7}
\end{equation}
where $d_{t_0}^{f}$ and $d_{t_1}^{f}$ represent the longitudinal distance between SV and the nearest vehicle in front before and after the driver model performs the action respectively. $\eta_2$ is the reward discount coefficient, which is set to 4000 in this paper.

The obtained cooperative rewards are shared by all driver models in the adversarial driving area. They are distributed to each model through the reward allocation mechanism, and are then added to the individual rewards, which are seen as the final reward for the current time-step action. In this study, the allocation mechanism of cooperation rewards is set to be evenly distributed to all driver models in the adversarial driving area. Finally, the reward value obtained by each model $i$ is:

\begin{equation}
\label{eq8}
    r_{i} = 
    \begin{cases} 
      r_{ind}, & \text{non-adversarial driving modality} \\ 
      r_{ind} + \frac{r_{coop}}{n}, & \text{adversarial driving modality}
    \end{cases}
\end{equation}
where $n$ is the number of BVs in the adversarial driving area.

\subsubsection{Behavioral Constraints} \label{behavConstraints}

In order to prevent the driver model from making meaningless dangerous driving behaviors that reduce the efficiency of scenario generation during simulation, some illegal and irrational behaviors of the model should be restricted. To achieve this, those behaviors are first identified and then awarded with a negative reward $r_{rv}$ during the individual reward calculation as described in Section \ref{sec_Reward}.

\begin{figure}[b] 
      \centering
      \includegraphics[width=.8\linewidth]{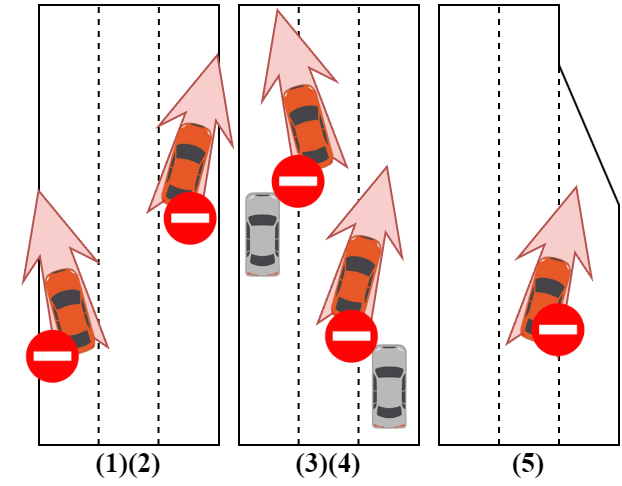}
      \caption{Behavioral constraints.}
      \label{BehavioralConstraint}
\end{figure} 

To be specific, behavioral constraints include five situations, as shown in Fig. \ref{BehavioralConstraint}: (1) When the vehicle is driving in the leftmost lane, changing lanes to the left is prohibited. (2) When the vehicle is driving in the rightmost lane, changing lanes to the right is prohibited. (3) When there are other vehicles within 10 meters in front and behind in the left lane of the vehicle, changing lanes to the left is prohibited. (4) When there are other vehicles within 10 meters in front and behind in the right lane of the vehicle, changing lanes to the right is prohibited. (5) Lane changes from the main road to merging ramps are prohibited. 

Among the behavioral constraints, constraint (1) and (2) are designed to prevent the driver model-controlled BVs from exiting road boundaries. And in constraints (3) and (4), we set a 10-meter front and rear buffer zone as the permissible lane-changing criterion. This configuration prevents BVs from either switching to congested lanes (which would degrade their travel efficiency)  or colliding with the SV through abrupt lane changes, thereby enabling the generation of more reasonable safety-critical scenarios. Constraint (5) prohibits BVs from entering merging ramps from the main road, which is considered irrational and may even violate traffic laws in certain countries.



\subsection{Dual-DM Training}
\subsubsection{Training Environment Settings}
The parameter settings of the driver model training environment are divided into simulation environment parameters and training algorithm parameters, as shown in Table \ref{tab2}. The simulation environment parameters set the number of vehicles of each type and the speed limits of vehicles in the simulated traffic environment. The training algorithm parameters set the relevant parameters of the RL algorithm during the model training process. When generating a vehicle entity, it is randomly selected from the three vehicle models in CARLA: Audi A2, Dodge Challenger, and Lincoln MKZ, and the associated dynamic models are used to update the status of each vehicle.

The Stackelberg model \cite{li2022gametheoretic} is used as the SUT during the adversarial driving modality training stage,  which can be replaced with any other model for the SUT-in-the-Loop training when testing other decision-making and planning systems.

\begin{table*}[b]
\renewcommand{\arraystretch}{1.3}
\caption{Parameter settings of Dual-DM training environment}
\label{tab2}
\begin{center}
\begin{tabular}{ m{0.13\linewidth} m{0.2\linewidth} m{0.35\linewidth} m{0.15\linewidth}<{\centering}}
\toprule
\textbf{Parameter type} & \multicolumn{2}{l}{\textbf{Parameter name}}& \textbf{Parameter setting} \\
\midrule
\multirow{5}{\linewidth}{Simulation environment parameters}  & \multirow{2}{\linewidth}{Non-adversarial driving modality training} & Number of trained model vehicles &  1  \\ \cline{3-4}
~ & ~ & Number of other vehicles in the environment & 12 \\ \cline{2-4}
~ & \multirow{2}{\linewidth}{Adversarial driving modality training} & Number of trained model vehicles & 12 \\ \cline{3-4}
~ & ~ & Number of SUT & 1 \\  \cline{2-4}
~ & \multicolumn{2}{l}{Maximum vehicle speed $v_{\text{max}} (\text{km/h})$}& Random in $\{90,100,110,120\}$ \\ \hline
\multirow{8}{\linewidth}{Training algorithm parameters} & \multicolumn{2}{l}{Training rounds}& 90000 \\ \cline{2-4}
~ & \multicolumn{2}{l}{Maximum number of replay buffer samples $\mathscr{B}_{\text{max}}$} & 40000 \\ \cline{2-4}
~ & \multicolumn{2}{l}{Strategic network learning rate $\alpha$} & 0.0001 \\ \cline{2-4}
~ & \multicolumn{2}{l}{Value function network learning rate $\beta_1, \beta_2 $} & 0.0001 \\ \cline{2-4}
~ & \multicolumn{2}{l}{Number of training samples per batch} & 10 \\ \cline{2-4}
~ & \multicolumn{2}{l}{Number of training iterations per round} & 2 \\ \cline{2-4}
~ & \multicolumn{2}{l}{Network soft update coefficient $\tau$} & 0.095 \\ \cline{2-4}
~ & \multicolumn{2}{l}{Discount coefficient $\gamma$} & 0.9 \\ 
\bottomrule
\end{tabular}
\end{center}
\end{table*}

\subsubsection{Convergence Curve of Reward Function}
\begin{figure}[b] 
      \centering
      \includegraphics[width=\linewidth]{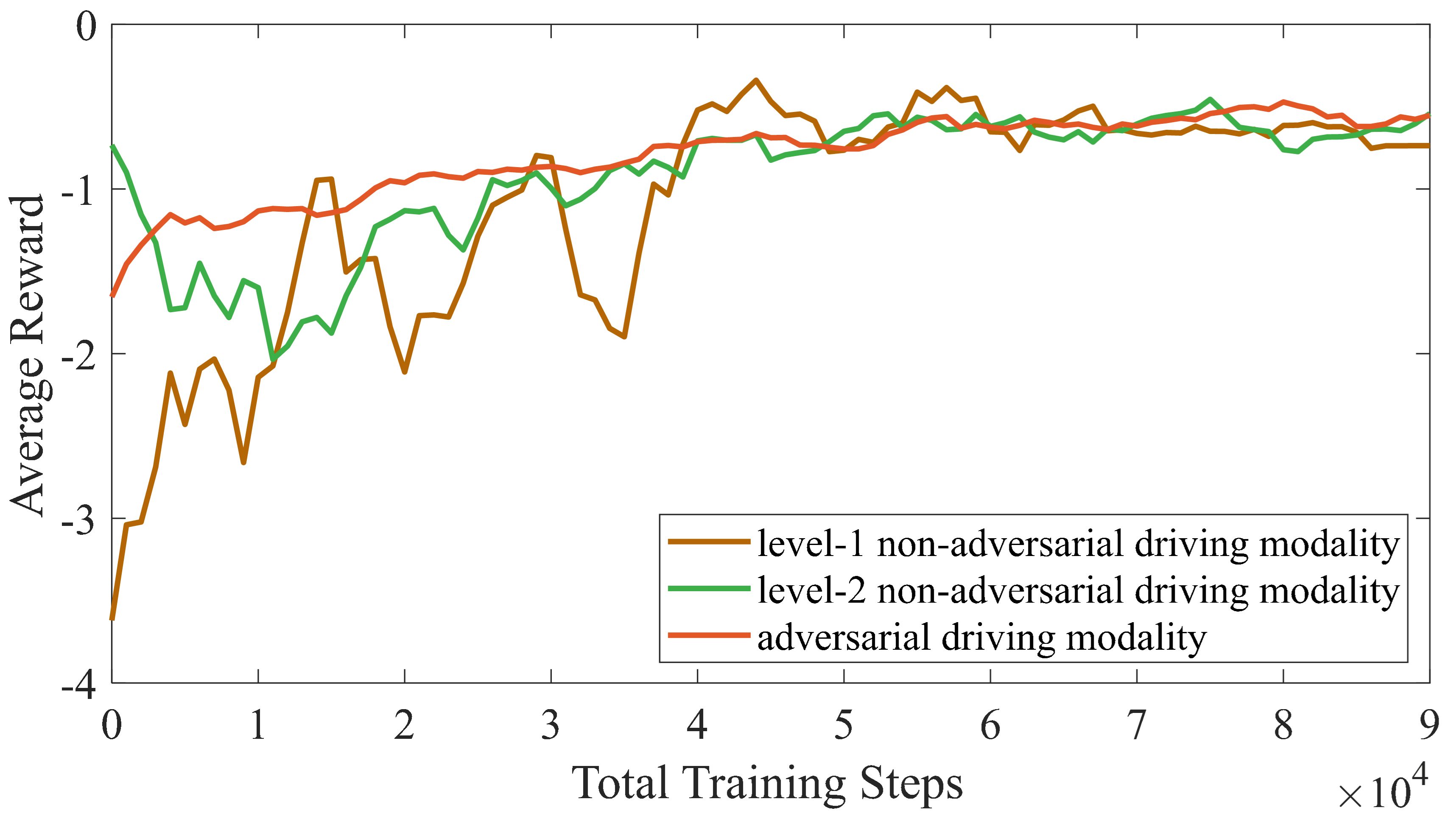}
      \caption{Dual-DM training reward curve.}
      \label{TrainingReward}
\end{figure}

Dual-DM is trained according to the designed training framework, and the average reward during the training process for different driving modalities is shown in Fig. \ref{TrainingReward}.

The average reward curves demonstrate that both level-1/level-2 non-adversarial driving modality training and adversarial driving modality training converge around 60,000 training rounds. Notably, the level-1 non-adversarial training initiates from a randomized model, causing its reward curve to originate at the minimum value before ascending. In contrast, both level-2 non-adversarial and adversarial training curves exhibit higher starting points, as their initial strategies inherit pre-trained capabilities from respective preceding stages (level-1 and level-2 non-adversarial training) rather than beginning with random parameters. Moreover, post-convergence analysis reveals marginally superior rewards in adversarial training because of the extra cooperative reward for adversarial driving modality.

\section{Evolving Scenarios Generation}
In this section, the well-trained Dual-DM is used as an NPC model to generate the behaviors of BVs in the continuous traffic simulation environment, interacting with SV controlled by SUT and continuously generation evolving scenarios. After the simulation, critical evolving scenarios are selected for subsequent analysis. Meanwhile, to validate the superiority of the proposed Dual-DM, comparative experiments are conducted using baseline driver models as NPCs to generate evolving scenarios.

\subsection{Simulation Map}
\begin{figure*}[b] 
      \centering
      \includegraphics[width=.9\linewidth]{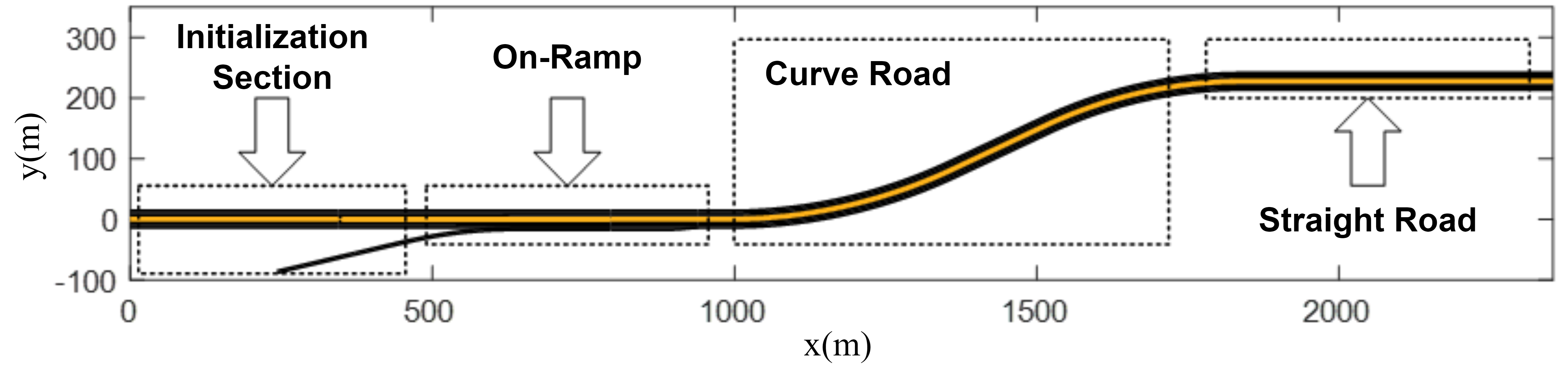}
      \caption{Simulation map.}
      \label{SimulationMap}
\end{figure*}

A highway map with four lanes is designed in this paper according to the requirements of highway autonomous driving testing, which includes straight lanes, curves and an on-ramp in road topology. The total length of the map is approximately 2400 meters. The road topology of the map is shown in Fig. \ref{SimulationMap}. The initialization area is used to allocate the position of the SV and the BVs and generate the vehicle entities. After the generation, the vehicles will go through the on-ramp, curves and straight lanes until they reach the road end.

\subsection{SUT and NPC models}
In the experiments, the Stackelberg model is chosen as the SUT to control the SV, which is consistent with the SUT selected during the training of Dual-DM. As for NPC models, besides the proposed Dual-DM, we additionally choose a function-based and a learning-based driver models respectively as baseline models for comparison. Two baseline models are elaborated as follows.

\subsubsection{Nilsson driver model (Nilsson)}
The Nilsson model is a typical lane-changing decision model based on predefined lane utility functions. This algorithm first determines whether a lane-changing maneuver is feasible. If feasible, it selects appropriate inter-vehicle traffic gaps and timing to execute the lane change, while calculating corresponding longitudinal and lateral control trajectories.

\subsubsection{DRL-based social driver model (Social-DRL)}
The DRL-based social driver model is trained and generated using the DRL method in our previous work \cite{ma2024evolving}. In fact, the training method used in the first stage of Dual-DM originates from Social-DRL. However, unlike the first-stage training of Dual-DM, Social-DRL incorporates additional social rewards in its reward function, thereby enabling the driver model to exhibit certain social attributes during interactions. Specifically, the model contains three social attributes: cooperative, neutral, and competitive. In the experiments of this study, the proportion of models is uniformly set to 30\% of cooperative, 30\% of neutral, and 40\% of competitive. 

\subsection{Signal Interface and Control Strategy}

\begin{figure}[t] 
      \centering
      \includegraphics[width=\linewidth]{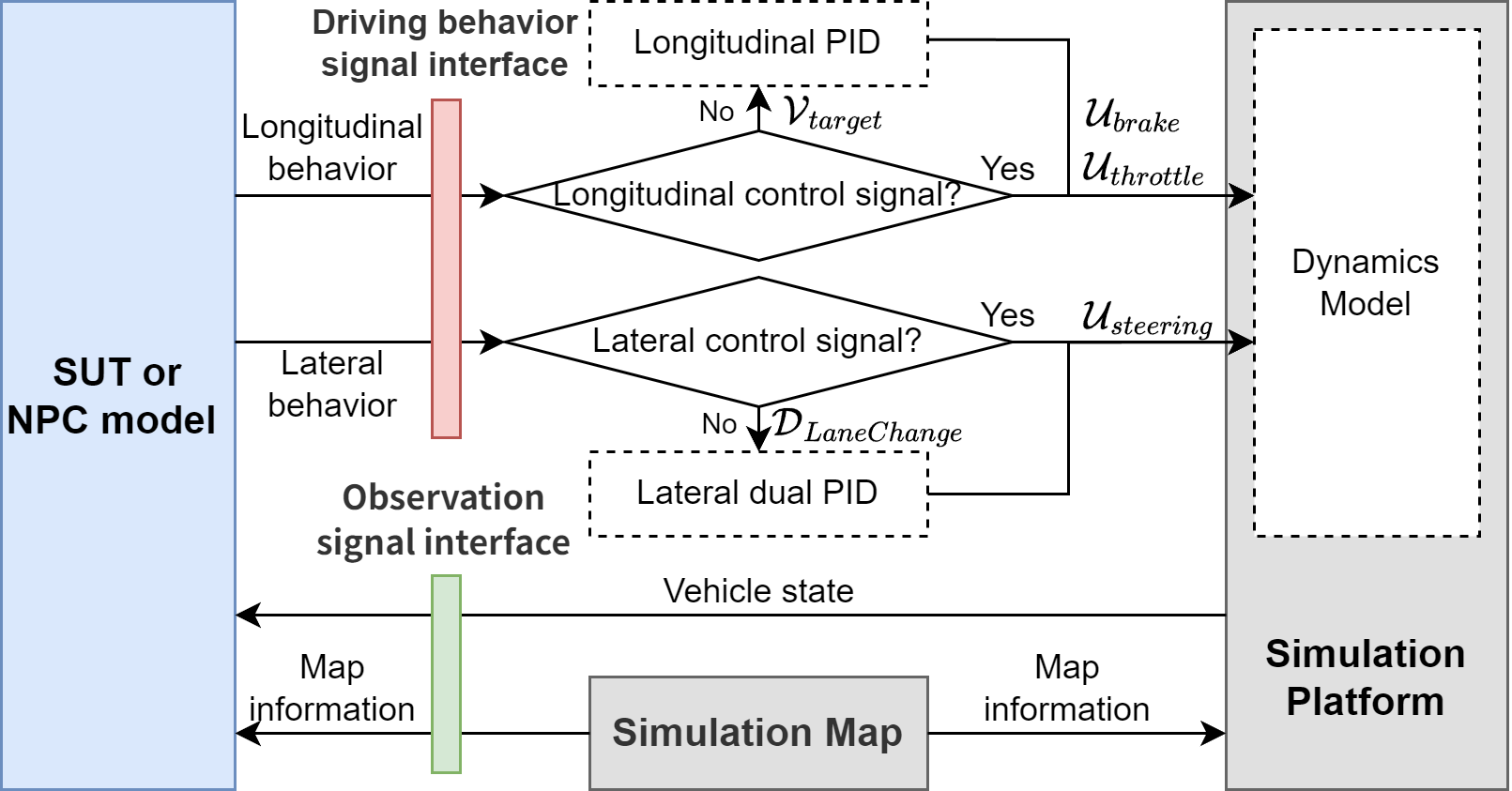}
      \caption{NPC model-simulation platform interface.}
      \label{Interface}
\end{figure}

To enable any decision-making and planning system (namely SUT) or driver models with such capabilities (namely NPC models) to be rapidly integrated into the simulation environment developed in this study, two interfaces, which transmit observation signals and driving behavior signals respectively, are designed, as shown in Fig. \ref{Interface}.

During the simulation, the SUT/NPC model makes behavioral decision based on environmental information through observation signal interfaces, including both map information and surrounding vehicle status information. And this information can be customized based on model requirements. Specifically, Map information is extracted through simulation map analysis, while vehicle status information is retrieved from the real-time simulation platform. At the same time, through the driving behavior signal interface, the NPC model/SUT will output longitudinal/lateral behavioral decisions or direct control signals to manipulate automotive entities in the simulation platform. 

\begin{table}[b]
\renewcommand{\arraystretch}{1.3}
\caption{The Output Control Signal Types of Different SUT/NPC Models}
\label{SignalType}
\begin{center}
\begin{tabular}{ m{0.12\linewidth} m{0.18\linewidth} m{0.27\linewidth}<{\centering} m{0.2\linewidth}<{\centering}}
\toprule
\textbf{Character} & \textbf{Model name} & \textbf{Longitudinal signal}  & \textbf{Lateral signal}\\
\midrule
SUT & Stackelberg & $\mathcal{V}_{target}$ & $\mathcal{D}_{LaneChange}$ \\ \hline
\multirow{3}{\linewidth}{NPC} & Nilsson  & $\mathcal{V}_{target}$ & \multirow{3}{\linewidth}{$\mathcal{D}_{LaneChange}$} \\
~ & Social-DRL & $\mathcal{U}_{throttle}\ \& \ \mathcal{U}_{brake}$   & ~ \\ 
~ & Dual-DM & $\mathcal{U}_{throttle}\ \& \ \mathcal{U}_{brake}$ & ~\\
\bottomrule
\end{tabular}
\end{center}
\end{table}

As for the control strategy, the outputs by SUT/NPC models can be strategic decisions (target speed $\mathcal{V}_{target}$ / lane change decision $\mathcal{D}_{LaneChange}$) or operational control signals (throttle $\mathcal{U}_{throttle}$ / brake $\mathcal{U}_{brake}$ / steering $\mathcal{U}_{steering}$), as shown in Table \ref{SignalType}. When operational control signals are transmitted, these signals are directly fed into the dynamic model of the simulation platform for control. Otherwise, when strategic decision-level outputs are transmitted, these instructions are translated into control signals via Proportion-Integration-Differentiation (PID) controllers before execution. More in detail, for longitudinal control, since the target speed generated by the driver model are continuous variables, they can be directly processed by a typical PID model. For lateral control, where the driver model outputs discrete lane change decision signal, a dual PID model is constructed with two objectives: (1) minimizing lateral deviation $d$ from the target lane centerline, and (2) minimizing the yaw angle discrepancy $e$ relative to the road geometry, thereby generating steering wheel angle control signals, as shown in Eq. \ref{eq9}.

\begin{equation}
\mathcal{U}_{steering} = \text{CLIP} [\alpha \text{PID}(d) + \beta \text{PID}(e), s_{min}, s_{max}]
\label{eq9}
\end{equation}
where $\alpha$ and $\beta$ are weighting coefficients used to balance the outputs of the two PID models, while $s_{min}$ and $s_{max}$ represent the vehicle's physical steering limits. The CLIP() function constrains the model's output within these physical steering limits. In this paper, $\alpha$ = 1.5 and $\beta$ = 1.0.

\subsection{Evolving Scenarios Generation and Critical Scenarios Filtering}

This paper selects CARLA as the basis for building the simulation platform. The simulation environment contains totally 16 vehicles, including 15 BVs controlled by a certain NPC model and 1 SV controlled by the SUT. The simulation will stop after 500 rounds. In each round, if a collision occurs between the SV and BV, between BVs, or if the SV reaches the end of the road, the simulation will restart for a new round. 

\begin{table}[b]
\renewcommand{\arraystretch}{1.3}
\caption{Safety-critical Scenario Generation Results after 500 Rounds of Simulation Experiments}
\label{tab3}
\begin{center}
\begin{tabular}{ m{0.21\linewidth} m{0.1\linewidth}<{\centering} m{0.001\linewidth}<{\centering} m{0.15\linewidth}<{\centering} m{0.15\linewidth}<{\centering} m{0.1\linewidth}<{\centering}}
\toprule
\multicolumn{2}{c}{\textbf{Model configuration}} & ~ & \multicolumn{3}{c}{\textbf{Number of safety-critical scenarios}} \\ \cline{1-2} \cline{4-6}
NPC model & SUT & ~ & Crash scenarios & Near-crash scenarios & Total \\ 
\midrule
Nilsson  & \multirow{3}{\linewidth}{Stackelberg} & ~ & 62 & 67 & 129  \\
Social-DRL & ~ &~ & 104 & 41 & 145 \\ 
Dual-DM (ours) & ~ &~ & 206 & 223 & 429 \\ 
\bottomrule
\end{tabular}
\end{center}
\end{table}

After completing all simulation experiments using different driver models as NPC models, safety-critical scenarios are identified and filtered from all recorded simulation data. Specifically, a safety-critical situation is first determined as either a collision between the SV and BVs (crash situation) or the SV's TTC being less than 0.5 seconds (near-crash situation). Next, trajectory data (including position, velocity, and acceleration information) of the SV and all BVs from the 3.5s preceding the safety-critical situation is extracted and saved as a safety-critical scenario. After 500 rounds of simulation experiments, the safety-critical scenarios generated through continuous interaction between various NPC models and the SUT are presented in Table \ref{tab3}. In addition to the total number of safety-critical scenarios generated by each NPC model, Table \ref{tab3} also provides the numbers of crash scenarios and near-crash scenarios.

\section{Evolving Scenarios Evaluation}
As previously mentioned, the goal of this paper is to efficiently generate complex and diverse critical evolving scenarios while guaranteeing scenario fidelity based on the trained Dual-DM. Therefore, this section evaluates the generated evolving scenarios in terms of fidelity, testing efficiency, complexity, and diversity. The evaluation framework is shown in the rightmost part of Fig. \ref{framework}. 

For the fidelity evaluation, all generated evolving scenarios are compared against naturalistic driving data, while only critical scenarios are utilized for comparative analysis in the other evaluation dimensions. Specifically, two metrics are developed to evaluate testing efficiency and complexity and the superiority is verified through comparisons with baseline models. For diversity evaluation, statistical analysis and case studies are conducted based on computational results of cooperative rewards.

\subsection{Fidelity Evaluation}
Fidelity aims to evaluate the similarity between naturalistic driving data and the simulation data of vehicles in evolving scenarios. The higher the fidelity of generated evolving scenarios, the greater their probability of real-world occurrence, thereby providing more significant guidance for further development of AVs. In this paper, The HighD dataset \cite{krajewski2018highda} is chosen as the benchmark for fidelity calculation. 

It is particularly noteworthy that despite the adversarial settings introduced in Dual-DM, only a limited number of BVs within the confined adversarial driving area operate in adversarial modality. The majority of BVs in the evolving scenarios maintain the natural, non-adversarial driving modality. Consequently, evaluating the fidelity of these evolving scenarios at the macroscopic traffic flow level remains statistically meaningful.

This paper employs Jensen-Shannon (JS) divergence as a refined metric for quantifying similarity of probability distributions between the generated evolving scenarios and HighD scenarios. Unlike the asymmetric limitation inherent in Kullback-Leibler (KL) divergence when characterizing distributional relationships, JS divergence provides a symmetrical measure through its entropy-based formulation. For two distinct probability distributions $P$ and $Q$, $JS(P||Q)$ quantifies the information-theoretic discrepancy between them by averaging the KL divergences relative to their mixture distribution. The JS divergence value decreases as the similarity between the distributions increases, reaching zero when the two distributions are identical. The formulation of $JS(P||Q)$ is expressed as:

\begin{equation}
   JS(P||Q) = \frac{1}{2} KL\left(P||M\right) + \frac{1}{2} KL\left(Q||M\right)
   \label{eq10}
\end{equation}
In which,

\begin{equation}
   M = \frac{1}{2}(P+Q)
   \label{eq11}
\end{equation}

\begin{equation}
   KL(G||M) = \sum_x g(x) \log \frac{g(x)}{m(x)}, G\in\{P,Q\}
   \label{eq12}
\end{equation}

\begin{figure}[!b]%
    \centering
    \subfloat[Probability distributions of velocity.]{
        \label{VelocityDis}
        \includegraphics[width=.9\linewidth]{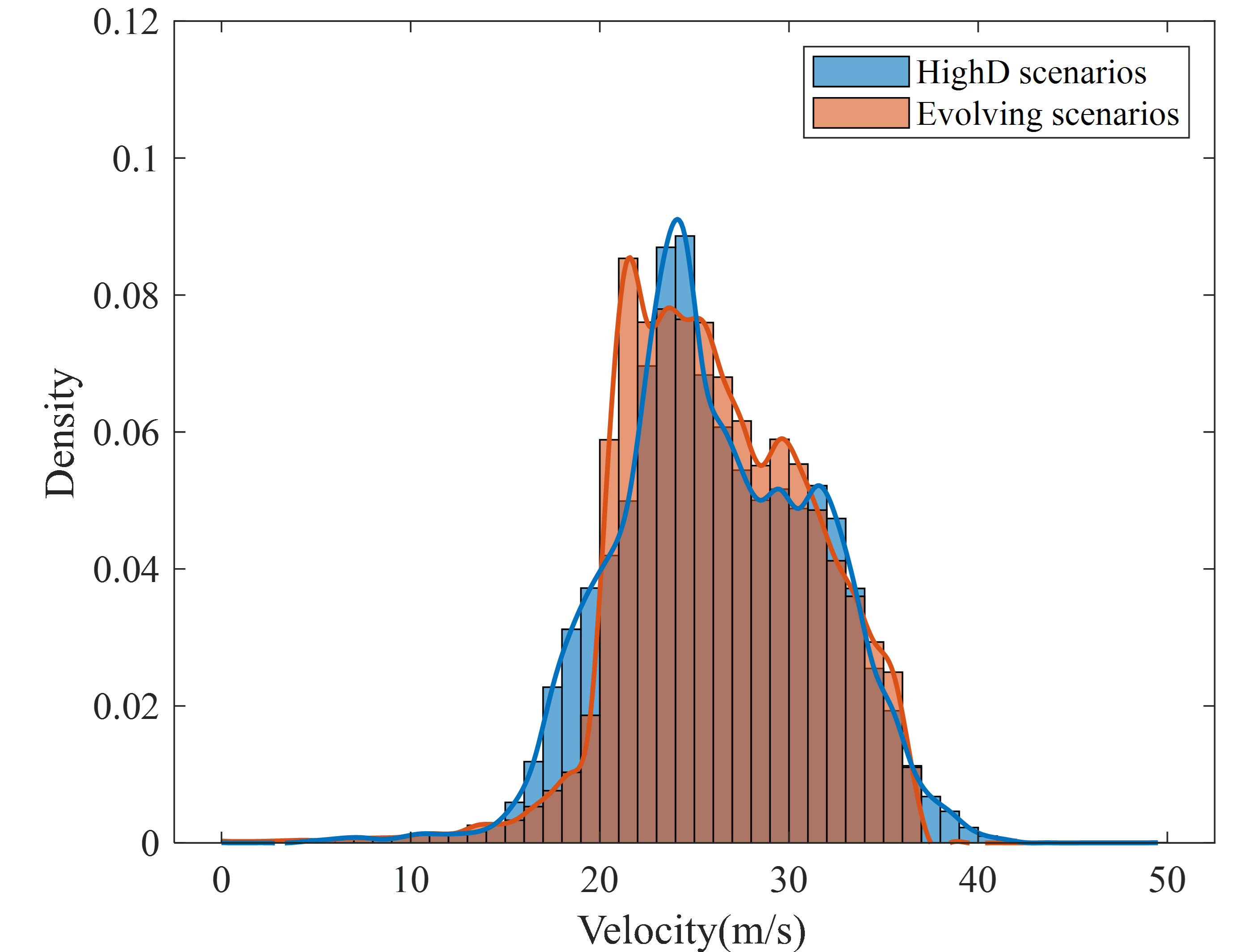}
        } \hfill
    \subfloat[Probability distributions of TTC under lane-change behaviors.]{
        \label{TTCDis}
        \includegraphics[width=.9\linewidth]{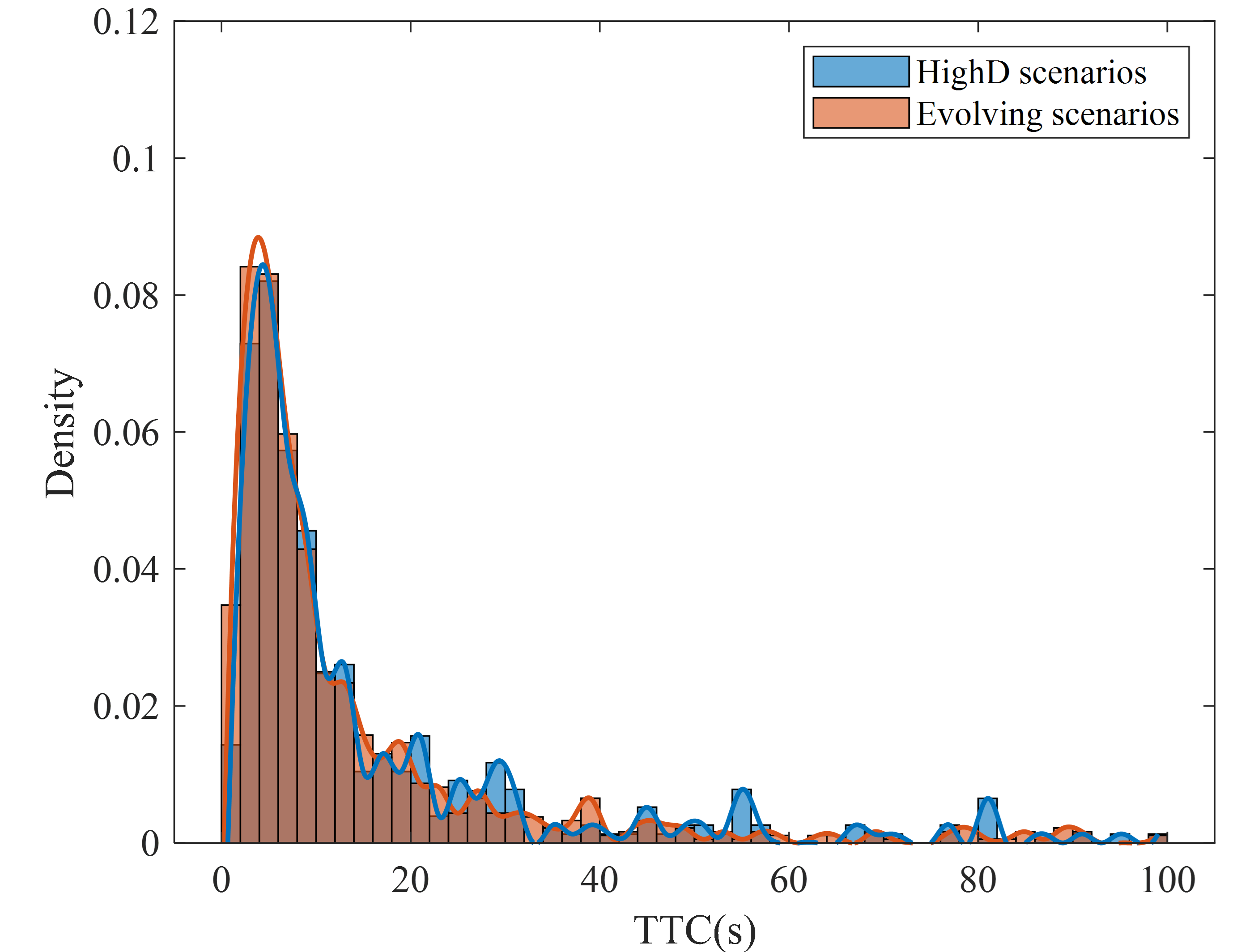}
        }
    \caption{Comparison of probability distributions for two behavioral metrics between HighD scenarios and the generated evolving scenarios.
    }
    \label{JScomparision}
\end{figure}

Given that the fidelity of Nilsson and Social-DRL has been thoroughly evaluated in their respective studies, this work focuses solely on evaluating the fidelity of evolving scenarios generated by Dual-DM. In this paper, the velocity of vehicles and the TTC when a vehicle performs a lane change maneuver are chosen as the comparative behavioral metrics. And their probability distributions in the HighD scenarios and all the evolving scenarios generated by BVs controlled by the Dual-DM are illustrated in Fig. \ref{JScomparision}. Thanks to the first-stage training of Dual-DM and the behavioral constraints designed in the individual reward function, Dual-DM demonstrates natural and realistic behaviors. As demonstrated in the figure, the probability distributions of both behavioral metrics for BVs controlled by Dual-DM closely align with those of real-world vehicles.

To further quantify this similarity, the JS divergence between the two types of scenarios on both behavioral metrics is calculated, and the results are shown in Table \ref{tab4}. The calculation results of two behavioral metrics are both smaller than 0.15, which indicates that the characteristics of vehicle behaviors in evolving scenarios are highly similar (more than 85\%) to those in naturalistic driving data. Consequently, the evolving scenarios can be regarded as high fidelity.

\begin{table}[t]
\renewcommand{\arraystretch}{1.3}
\caption{JS Divergence between Data Distributions from Evolving Scenarios and HighD Scenarios.}
\label{tab4}
\begin{center}
\begin{tabular}{ m{0.5\linewidth} m{0.3\linewidth}<{\centering} }
\toprule
\textbf{Metrics} & $JS (HighD||Evolving)$  \\
\midrule
Velocity  & 0.0074 \\
TTC under lane-change behaviors & 0.1323\\ 
\bottomrule
\end{tabular}
\end{center}
\end{table}

\subsection{Test Efficiency Evaluation}
Test efficiency measures the speed of generating safety-critical scenarios. The more safety-critical scenarios generated within limited simulation rounds, the more likely it is to expose flaws in the decision-making and planning system, and also higher practical application value of the driver model. The efficiency is quantified by the number of safety-critical scenarios generated during the continuous interaction between the driver model and the SUT in a unit number of simulation rounds. Therefore, the calculation formula of the efficiency evaluation result $E$ is:

\begin{equation}
   E = 
   \begin{cases} 
   \frac{n}{N}, &  \frac{n}{N} \leq 1 \\
   1, &  \frac{n}{N} > 1 
   \end{cases}
\label{eq13}
\end{equation}
where $N$ is the number of simulation rounds, $n$ is the number of critical scenarios generated in $N$ rounds of simulation. Notably, since a single simulation round can generate multiple safety-critical scenarios (e.g., after recovering from a near-crash situation, the SV resumes normal driving but subsequently collides with other BVs), the number of generated safety-critical scenarios may exceed the total number of simulation runs. In such cases, we consider the safety-critical scenario generation efficiency to be 100\% as shown in Eq. \ref{eq13}.

The comparison of the efficiency evaluation results is shown in Table \ref{tab5}. In the efficiency evaluation, when Dual-DM is used as the NPC model, 429 safety-critical scenarios are generated in 500 rounds of testing, which is the largest number among the three NPC model settings, resulting in the efficiency evaluation results reaching 0.86, which is more than 195\% higher than the two baseline NPC models. As for the two baseline models, since they are both dedicated to generating vehicle behaviors that better conform to the naturalistic distribution, the number of critical scenarios generated is only 129 and 145 respectively. 

\begin{table}[t]
\renewcommand{\arraystretch}{1.3}
\caption{Comparison of Test Efficiency Evaluation Results}
\label{tab5}
\begin{center}
\begin{tabular}{ m{0.25\linewidth} m{0.25\linewidth}<{\centering} m{0.3\linewidth}<{\centering}}
\toprule
\textbf{NPC model} & \textbf{Total number of critical scenarios} & \textbf{Efficiency evaluation results $E$}  \\
\midrule
Nilsson  & 129 & 0.26 \\
Social-DRL & 145 & 0.29 \\ 
Dual-DM (ours) & \textbf{429} & \textbf{0.86}\\
\bottomrule
\end{tabular}
\end{center}
\end{table}

In particular, when compared to Social-DRL, which can be viewed as our previous work where a driver model operates only in non-adversarial modality, Dual-DM achieves a substantial enhancement in the efficiency of generating safety-critical scenarios without performance degradation in scenario fidelity and complexity. This demonstrates the effectiveness of the multi-vehicle cooperative adversarial configuration in enhancing the efficiency of safety-critical scenario generation, which can greatly improve the probability of evolving into safety-critical scenarios through continuous adversarial interaction against SV.

\subsection{Complexity Evaluation}
Scenario complexity is a quantitative metric to evaluate the challenge level of the generated evolving scenarios. The higher the complexity of generated evolving scenarios, the more challenging they become for the SUT, thereby a higher value for testing. In this study, the complexity of a generated scenario is quantified through two aspects: the number and distribution of BVs involved and the complexity of the BVs' actions in the scenario. 

For the first aspect, we introduce the concept of information entropy to quantify scenario complexity $C_{pos}$. Information entropy measures the average uncertainty of a random variable. Applied to driving scenarios, when there are more BVs with greater spatial dispersion, the scenario exhibits higher uncertainty and greater information entropy, corresponding to increased complexity. To obtain  $C_{pos}^i$ of a certain scenario $i$ among all the generated critical scenarios, the space near SV in the scenario is first divided into 8 areas as shown in Fig. \ref{ComplexityCal}, and $C_{pos}^i$ can be calculated as follows:

\begin{figure}[b] 
      \centering
      \includegraphics[width=.9\linewidth]{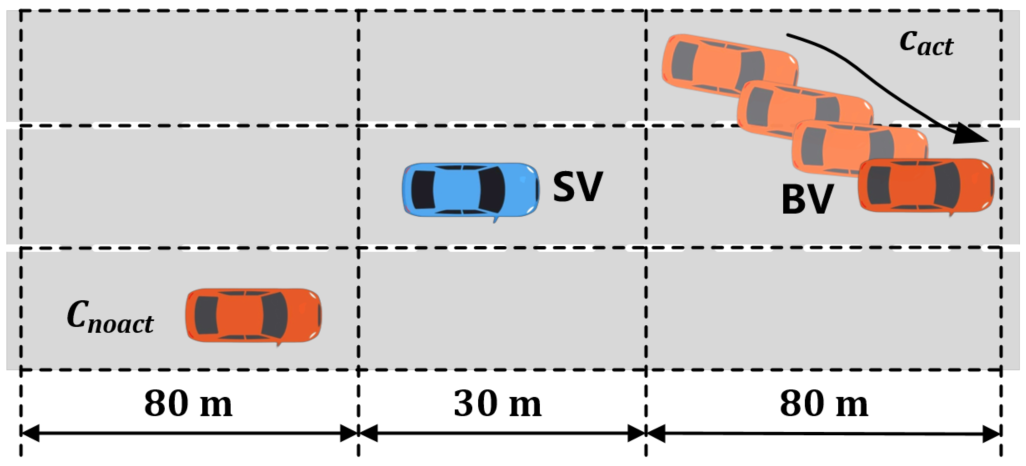}
      \caption{Complexity calculation of generated evolving scenarios.}
      \label{ComplexityCal}
\end{figure}

\begin{equation}
   C_{pos}^i = 
   \begin{cases} 
   -\sum_{j=1}^8 p(x_{j})\text{log}_2p(x_{j}), &  p(x_j) \neq 1 \\
   0.5, &  p(x_j) = 1 
   \end{cases}
\label{eq14}
\end{equation}
in which,
\begin{equation}
   p(x_j) = \frac{n_j}{N} 
\label{eq15}
\end{equation}
\begin{equation}
   N=\sum_{j=1}^8 n_j
\label{eq16}
\end{equation}
where $j$ denotes the $j$-th divided area, and $n_j$ represents the number of BVs in the $j$-th area.

\begin{figure}[b] 
      \centering
      \includegraphics[width=.9\linewidth]{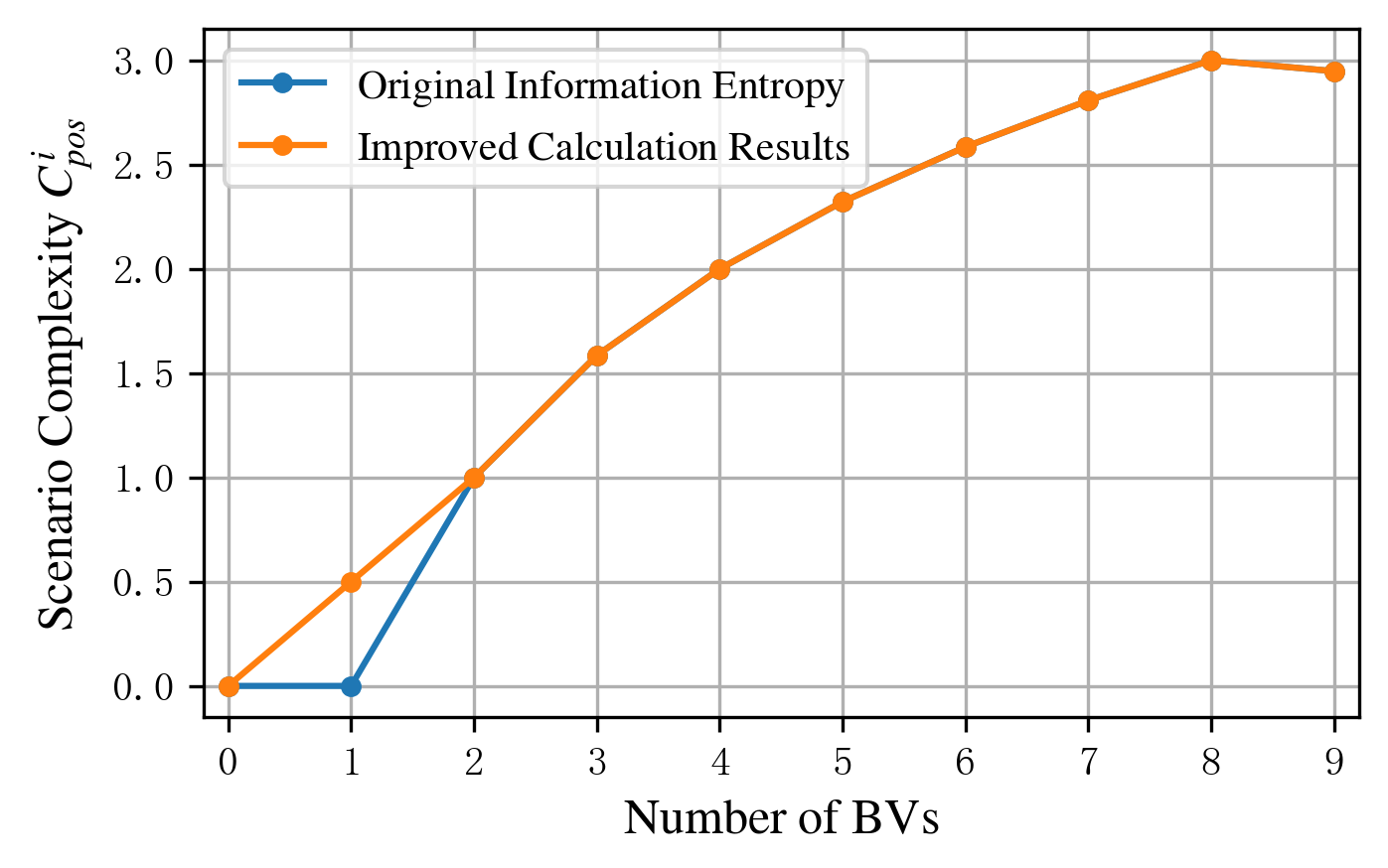}
      \caption{Illustration of the $C_{pos}^i$ calculation method.}
      \label{ComplexityRes}
\end{figure}

To clearly illustrate the calculation method for $C_{pos}^i$, in Fig. \ref{ComplexityRes} we compute scenarios with 1 to 9 BVs surrounding the SV according to Eq. \ref{eq14}. As the number of BVs increases, newly added BVs are preferentially placed in areas without existing BVs. When there are 9 BVs, one specific area contains two BVs while all others maintain single-vehicle occupancy. As depicted in Fig. \ref{ComplexityRes}, the maximum value $C_{pos}^{i}=3$ is achieved when all eight surrounding areas of the SV contain one BV. Then $C_{pos}^i$ decreases with nine BVs due to the relative concentration of vehicles (two BVs located in the same area).
Notably, if all BVs in the scenario are concentrated in a single area (i.e., $p(x_j) = 1$), the original information entropy calculation yields $C_{pos}^{i}=0$, which contradicts common sense. To address this, we assign $C_{pos}^{i}=0.5$ when $p(x_j) = 1$. A comparison of the calculation results before and after the improvement is also presented in Fig. \ref{ComplexityRes}.

For the second aspect of complexity evaluation, we define the action complexity as $c_{act}$ when vehicles exhibit either acceleration/deceleration (with absolute acceleration $\ge 4m/s^2$), or lane-changing behaviors (identified by changes in lane ID), and as $c_{noact}$ for other situations. The overall action-induced complexity $C_{act}^i$ is then calculated as:

\begin{equation}
   C_{act}^i=\frac{n_{act}}{N}\times c_{act} + \frac{n_{noact}}{N}\times c_{noact}
\label{eq17}
\end{equation}
where $n_{act}$ and $n_{noact}$ represent the number of BVs with and without driving actions in the scenario, respectively. In this paper, we set $c_{act}=1$ and $c_{noact}=0$.

According to Eq. \ref{eq14} and Eq. \ref{eq17}, the complexity evaluation result for the $i$-th scenario is:
\begin{equation}
   C^i=\frac{C_{pos}^i \times C_{act}^i}{\eta_c}
\label{eq18}
\end{equation}
where $\eta_c$ is a normalization factor. Under the above-mentioned settings, the theoretical maximum value of the complexity evaluation result $C^i$ for an evolving scenario is 3, therefore we set it to $\eta_c$.

Finally, the complexity of all the critical evolving scenarios generated by a certain driver model can be calculated as shown in Eq. \ref{eq19}.
\begin{equation}
   C_{DriverModel} = \frac{1}{m} \sum_{i=1}^{m} C^i
\label{eq19}
\end{equation}
where $m$ is the number of generated critical evolving scenarios by a certain driver model.

\begin{table}[t]
\renewcommand{\arraystretch}{1.3}
\caption{Comparison of Complexity Evaluation Results}
\label{tab6}
\begin{center}
\begin{tabular}{ m{0.25\linewidth} m{0.5\linewidth}<{\centering}}
\toprule
\textbf{NPC model} &  \textbf{Complexity evaluation results $C$}  \\
\midrule
Nilsson  & 0.34 \\
Social-DRL &  0.40 \\ 
Dual-DM (ours) &  \textbf{0.45}\\
\bottomrule
\end{tabular}
\end{center}
\end{table}

According to Eq. \ref{eq14} - \ref{eq19}, the complexity of critical evolving scenarios generated by different NPC models is obtained, as shown in Table \ref{tab6}. Among the three NPC models, Dual-DM performs the best in terms of complexity, reaching a complexity score of 0.45, which is 32.35\% and 12.5\% higher than Nilsson and Social-DRL respectively.

When it comes to the Social-DRL, its complexity result appears to show no significant gap compared to Dual-DM. We attribute this to the fact that complexity is calculated exclusively in safety-critical scenarios. Once a scenario is identified as safety-critical, it inherently presents challenges, leading to naturally higher complexity metrics. However, when comparing scenario complexity alongside the efficiency of safety-critical scenario generation, a substantial disparity in scenario generation performance between Dual-DM and Social-DRL will become evident.

\subsection{Diversity Evaluation}

The three evaluation dimensions discussed above assess the generated evolving scenarios based on external scenario information (i.e., vehicle trajectories). This section will further investigate the diversity of evolving scenarios generated by Dual-DM by incorporating internal model information (i.e., the model's reward function) to identify BVs' adversarial behaviors. 
This paper defines diversity as the richness of adversarial interaction patterns between BVs and the SV within safety-critical evolving scenarios.

Since the baseline NPC models lack adversarial settings, their driving intentions cannot be determined through either internal model information or external scenario observations (even when a BV is in close proximity to the SV, it remains impossible to determine whether the BV intends to drive adversarially or has simply arrived near the SV through natural driving behaviors). Consequently, diversity evaluation is conducted exclusively through statistical analysis and case studies.

\subsubsection{Statistical Analysis}

As described in Section \ref{sec_Reward}, when BVs' cooperative reward values are positive, it indicates that all BVs within the adversarial driving area have effectively executed adversarial interactions with the SV. Specifically, this means the BVs have successfully compressed either the SV's forward drivable space or potential lateral lane-changing space. Building upon this insight, the diversity of critical evolving scenarios generated by Dual-DM can be investigated from two perspectives: 1) the number of BVs participating in adversarial interactions and 2) the position combinations of cooperative adversarial behaviors. 

\begin{figure}[b] 
      \centering
      \includegraphics[width=.9\linewidth]{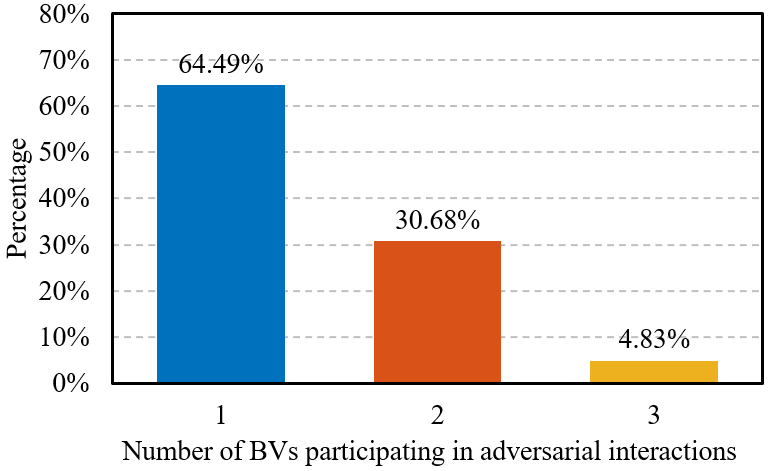}
      \caption{Distribution of the number of BVs participating in adversarial interactions.}
      \label{DistributionBVsNum}
\end{figure}

When analyzing the first aspect, the number of BVs participating in adversarial interactions in each scenario is counted, and the results are shown in Fig. \ref{DistributionBVsNum}. Among the 429 safety-critical evolving scenarios generated by Dual-DM, BVs receive positive cooperative rewards in 414 scenarios. As shown in Fig. \ref{DistributionBVsNum}, the number of BVs participating in adversarial interactions varies from 1 to 3 across scenarios generated with a positive cooperative reward, demonstrating the diversity of adversarial driving in these scenarios. As for the distribution of the number of BVs, the most common situation is that only 1 vehicle exists in the adversarial driving area, accounting for 64.49\%. The cases with 2 vehicles and 3 vehicles are 30.68\% and 4.83\% respectively. It can be clearly observed that the scenario proportion decreases as the number of BVs in the adversarial driving area increases. This is because the size of the adversarial driving area is limited, and more vehicles will lead to excessive local vehicle density. What’s more, from the perspective of the effect of adversarial behaviors, the compression of the drivable space in front of SV and lane-changing space can be well achieved by 3 vehicles. Therefore, there are usually no more than 3 BVs.

\begin{figure}[t] 
      \centering
      \includegraphics[width=.9\linewidth]{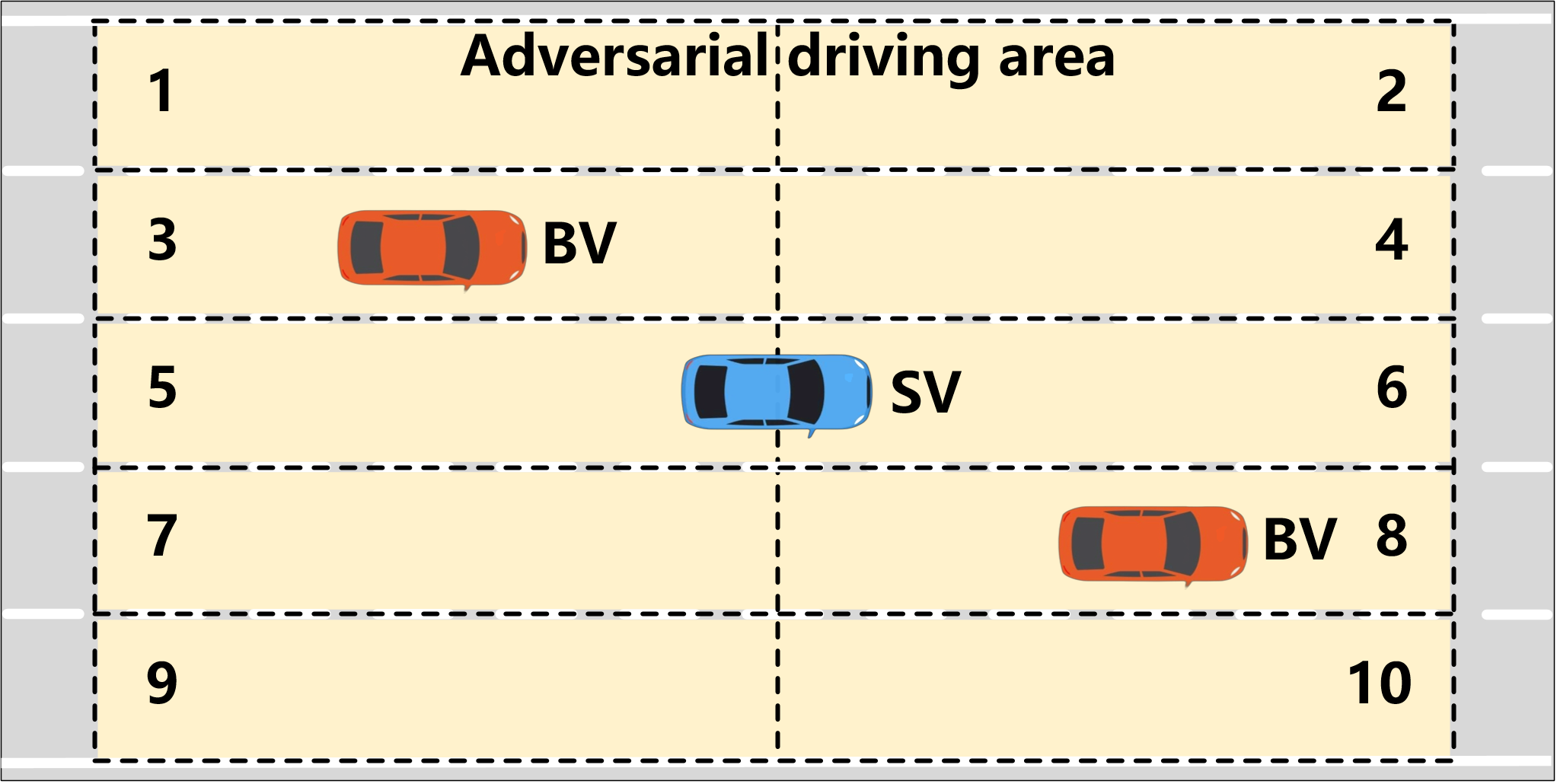}
      \caption{Division method of the adversarial driving area.}
      \label{DrivingPatterns}
\end{figure}

When it comes to the second aspect, the combination of the BV positions is analyzed. More in detail, the adversarial driving area is divided into 5 sections laterally according to the lanes, and each section is divided into 2 zones according to the longitudinal relationship relative to the SV, creating a total of 10 areas, as shown by the black dotted line in Fig. \ref{DrivingPatterns}. The combination of areas where the BVs start to perform effective adversarial behaviors is defined as the combined state of the BV positions in the current scenario. For instance, in Fig. \ref{DrivingPatterns}, there are two BVs around SV, which are located in Area 3 and Area 8 respectively. If these two BVs start to perform effective adversarial behaviors at this moment (i.e., these two BVs start to obtain positive cooperative rewards), the combined state of the BV positions for this scenario is [3, 8].

Scenarios with two and three BVs in the adversarial driving area are analyzed respectively. The results reveal that among 414 scenarios where BVs earn positive cooperative rewards, 127 two-BV cooperative adversarial scenarios demonstrate 23 distinct position combinations (namely 23 adversarial patterns), with 20 three-BV scenarios demonstrating 16 adversarial patterns. Therefore, the proposed driver model also demonstrates diversity in terms of the patterns of adversarial behaviors.

\subsubsection{Case Studies}

\begin{figure}[b] 
      \centering
      \includegraphics[width=\linewidth]{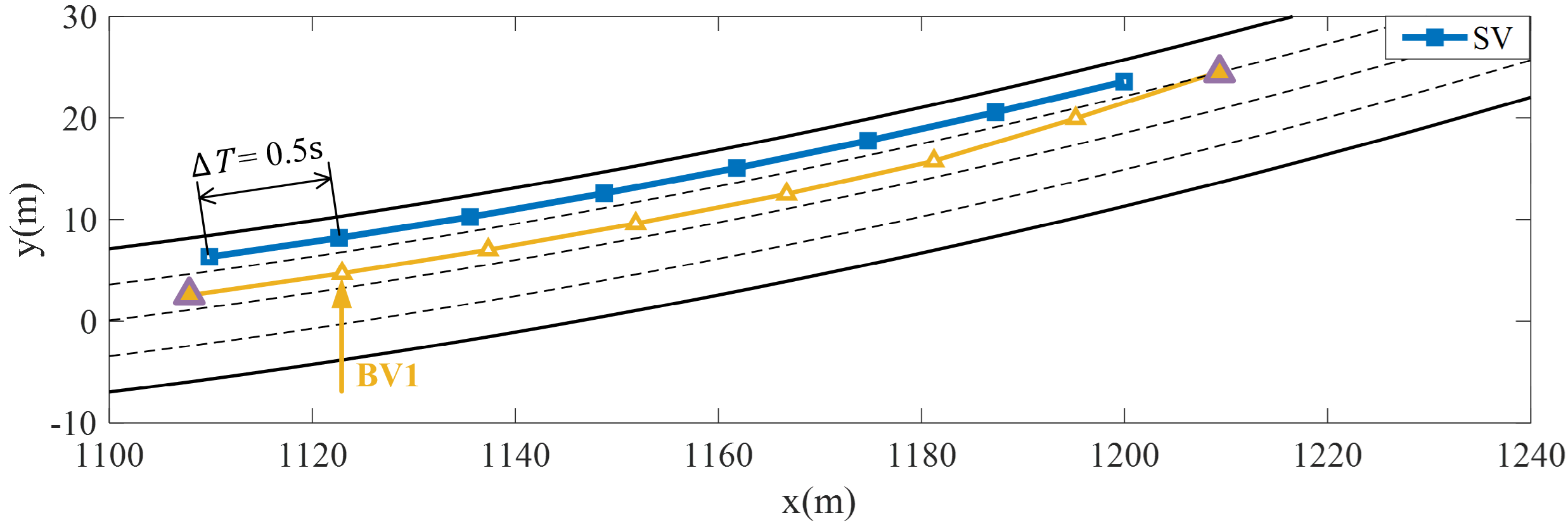}
      \caption{Visualization of single-vehicle adversarial behaviors.}
      \label{OneBV}
\end{figure} 

\begin{figure}[b] 
      \centering
      \includegraphics[width=\linewidth]{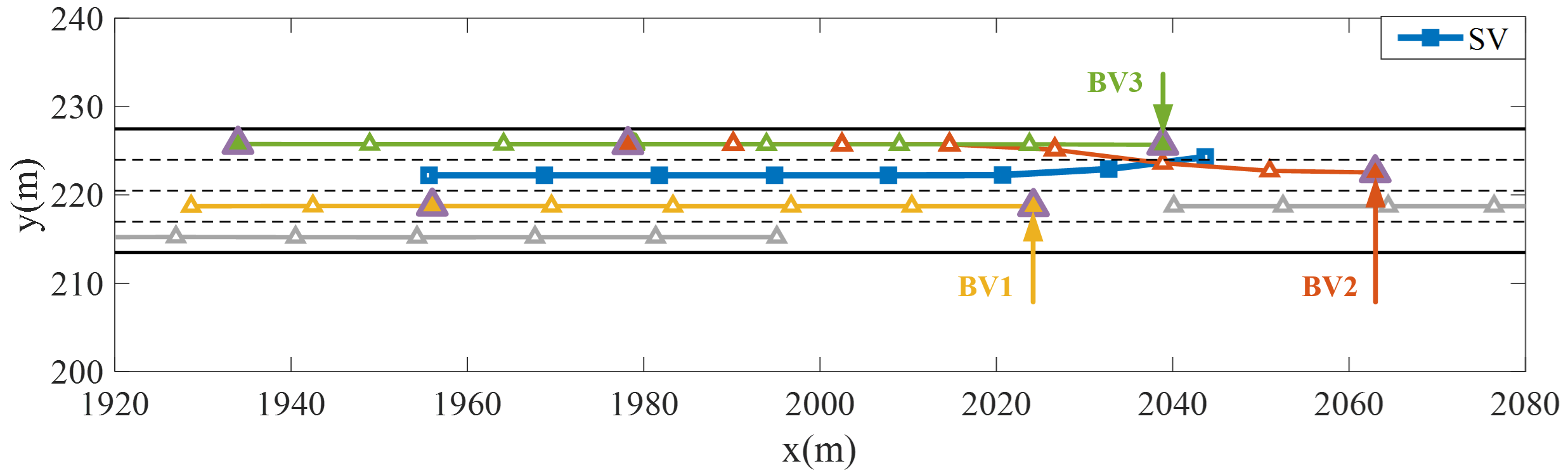}
      \caption{Visualization of three-vehicle cooperative adversarial behaviors.}
      \label{ThreeBV}
\end{figure}

Fig. \ref{OneBV}$\sim$\ref{ThreeBV} illustrate visualized scenario fragments of the Dual-DM-controlled BV(s) interacting with the SV, demonstrating both single-BV adversarial behaviors and multi-BVs cooperative adversarial behaviors involving two and three coordinated BVs. In these figures, the trajectory of the SV is indicated by a blue line, with blue squares marking the positions of the center of SV at each time step, while BVs displaying the adversarial driving behaviors are indicated by triangles, with a purple triangular edge representing when the BV enters or exits the adversarial driving area. Other BVs in the non-adversarial modality are indicated by grey lines and triangles. The simulation timestep $\Delta t = 0.1 \text{s}$, and each scenario lasts for 3.5s. For ease of visualization, we mark the vehicle's position every 5 timesteps (namely $\Delta T=0.5$s as shown in Fig. \ref{OneBV}). Moreover, all vehicles drive from left to right, with the leftmost lane in the driving direction designated as Lane 1, followed sequentially by Lanes 2, 3, and 4 to the right. 

Fig. \ref{OneBV} shows the Dual-DM-controlled BV confronting the SV in the case of a single vehicle. At first, SV controlled by SUT is driving in Lane 1, and BV1 is driving in Lane 2 behind the SV, the speed of which is higher than that of SV. Initially, BV1 maintains a high speed and drives in Lane 2, thereby compressing the SV's right lane-changing space. When BV1 reaches the front-right side of the SV, it executes a lane-changing maneuver to the left, thereby  further compressing the SV's forward drivable space. Through the above process, it can be observed that the proposed driver model possesses single-vehicle adversarial driving capability.

\begin{figure}[t] 
      \centering
      \includegraphics[width=\linewidth]{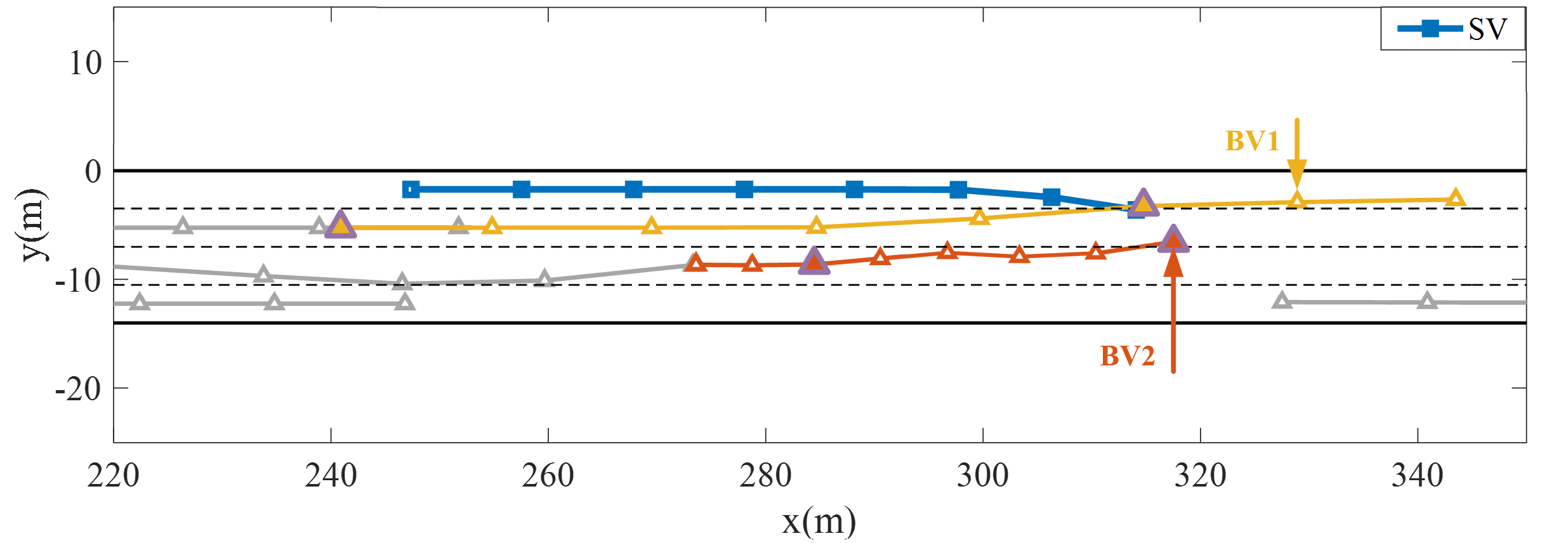}
      \caption{Visualization of two-vehicle cooperative adversarial behaviors.}
      \label{TwoBV}
\end{figure}

Fig. \ref{TwoBV} shows the adversarial behaviors of BVs when two vehicles cooperate with each other. In the beginning, SV controlled by SUT drives in Lane 1. Similar to the BV in Fig. \ref{OneBV}, BV1 drives at a high speed behind the SV, compressing its right lane-changing space. At $t =$1.5s, BV2 in Lane 3 enters the adversarial driving area. At this point, both BVs begin cooperative adversarial maneuvers against the SV: BV1 executes a left lane change ahead of the SV, reducing its forward drivable space, while BV2 simultaneously initiates a left lane change to replace BV1 in compressing the SV's right lane-changing space. As a consequence, SV attempts to avoid the conflict by changing lanes to Lane 2 after BV1's maneuver, but collides with BV2 during its ongoing leftward lane change. As shown in the figure, at $t=$3s, BV2 crosses the lane marking earlier, yet the SV persists in completing its lane change, ultimately resulting in a collision. This scenario demonstrates successful adversarial cooperation between the two BVs and reveals deficiencies in the SV’s decision-making and planning.

Fig. \ref{ThreeBV} shows the cooperative adversarial behaviors of BVs among three vehicles. Initially, SV controlled by SUT drives in Lane 2. BV1 and BV3 drive behind SV in Lane 3 and Lane 1 respectively. BV2 drives in front of BV3 in Lane 1. When the scenario starts, BV1 and BV3 maintain high speeds and drive in their respective lanes, progressively constraining the SV's lane-changing space on both sides. At $t=$2.5s, BV2 executes a right lane change that critically restricts the SV's forward drivable space. This maneuver forces the SV to initiate an evasive left lane change while ignoring BV2 approaching at high velocity from the rear, finally leading to a collision. It is worth noting that although the collision is caused by BV3 colliding with SV, at $t=$3s when the SV exhibits a clear lane-changing intention, the longitudinal distance between BV3's front and the SV's rear is less than 5m. For BV3, the longitudinal Time-to-Collision (TTC) is below 1 second at this moment, making collision avoidance practically impossible. Therefore, the primary cause of this incident remains attributable to deficiencies in the SV's decision-making and planning (specifically, its reckless lane-changing behavior).

From the three aforementioned cases, it is evident that the BVs challenge the SV through progressive compression of its drivable space rather than initiating direct collisions, which aligns with our design objectives, thereby qualitatively validating the effectiveness of Dual-DM. At the same time, the three cases further illustrate the diversity of the generated critical evolving scenarios in terms of the number of BVs participating in adversarial interactions.

\section{Conclusion}

In this paper, a dual-modal driver model (Dual-DM) with non-adversarial and adversarial driving modalities is designed and trained in order to generate safety-critical scenarios through evolving scenario generation method. First, Dual-DM along with its training framework and detailed training configurations are designed. Next, Dual-DM is trained and connected to the experimental environment to automatically generate safety-critical scenarios. After that, an evaluation framework for evolving scenarios are proposed. The fidelity, test efficiency, complexity and diversity of the Dual-DM-generated evolving scenarios are verified through comparisons with naturalistic scenarios, evolving scenarios generated by baseline models, as well as statistical analysis and case studies. 

Experimental results show that Dual-DM proposed in this paper can generate complex, diverse, and strong interactive safety-critical scenarios. Compared to other baseline models, Dual-DM achieves a substantial improvement in the efficiency of generating safety-critical scenarios without compromising scenario fidelity and complexity. The statistical analysis and case studies further demonstrate the diversity of safety-critical evolving scenarios generated by Dual-DM in terms of the number of adversarial BVs and their position combinations of cooperative adversarial interactions. These results confirm Dual-DM's capability in generating safety-critical scenarios and show Dual-DM's potential in future testing and evaluation for the decision-making and planning system of highly automated vehicles.

Future work will primarily focus on training the Dual-DM in more diverse road topologies to enhance its generalizability and scenario adaptability. Additionally, more advanced and widely adopted autonomous driving systems will be incorporated into the proposed testing framework as SUTs to better demonstrate the practical value of Dual-DM.



\bibliographystyle{IEEEtran}
\bibliography{IEEEabrv,citelist}


 





\end{document}